\documentclass[journal]{IEEEtran} 
\usepackage{indentfirst}
\usepackage{lineno}
\usepackage{arydshln}
\usepackage{booktabs}
\usepackage{colortbl}
\usepackage[table]{xcolor}
\usepackage{graphicx}
\usepackage{subfigure}
\usepackage{multirow}
\usepackage{longtable}
\usepackage{algorithm}
\usepackage{algorithmic}
\usepackage{stfloats}
\usepackage{caption2}
\usepackage{amssymb,amsmath}
\usepackage{epstopdf}
\usepackage{color}
\usepackage{rotating}
\usepackage{multirow}
\usepackage{threeparttable}
\usepackage{bbding}
\usepackage{comment}
\usepackage{cite}
\usepackage[colorlinks,linkcolor=red,anchorcolor=blue,citecolor=blue]{hyperref}
\definecolor{rrr}{rgb}{1    0	0.0392}
\definecolor{myred}{rgb}{0.7059    0	0.0392}					
\definecolor{myblue}{rgb}{0    0	0.9392}					

\ifCLASSINFOpdf
\else
\fi

\begin{document}
\title{United Domain Cognition Network for Salient Object Detection in Optical Remote Sensing Images}
\author{
        Yanguang~Sun,
		Jian~Yang,
        Lei~Luo
\thanks{This work was supported by the National Natural Science Foundation of China (Grant No. 62276135, 61806094, and 62176124). (\textit{Corresponding author: Lei Luo,} e-mail: luoleipitt@gmail.com)
	
Y. Sun, L. Luo, and J. Yang are with the PCA Laboratory, Key Laboratory of Intelligent Perception and Systems for High-Dimensional Information of Ministry of Education, School of Computer Science and Engineering, Nanjing University of Science and Technology, Nanjing, China. (e-mail: Sunyg@njust.edu.cn; luoleipitt@gmail.com; csjyang@mail.njust.edu.cn).

}
}

\markboth{Journal of \LaTeX\ Class Files}%
{Shell \MakeLowercase{\textit{et al.}}: Bare Demo of IEEEtran.cls for IEEE Journals}

\maketitle
\begin{abstract}
Recently, deep learning-based salient object detection (SOD) in optical remote sensing images (ORSIs) have achieved significant breakthroughs. 
We observe that existing ORSIs-SOD methods consistently center around optimizing pixel features in the spatial domain, progressively distinguishing between backgrounds and objects. However, pixel information represents local attributes, which are often correlated with their surrounding context. Even with strategies expanding the local region, spatial features remain biased towards local characteristics, lacking the ability of global perception. To address this problem, we introduce the Fourier transform that generate global frequency features and achieve an image-size receptive field. To be specific, we propose a  novel United Domain Cognition Network (UDCNet) to jointly explore the global-local information in the frequency and spatial domains. Technically, we first design a frequency-spatial domain transformer block that mutually amalgamates the complementary local spatial and global frequency features to strength the capability of initial input features. Furthermore, a dense semantic excavation module is constructed to capture higher-level semantic for guiding the positioning of remote sensing objects. Finally, we devise a dual-branch joint optimization decoder that applies the saliency and edge branches to generate high-quality representations for predicting salient objects. Experimental results demonstrate the superiority of the proposed UDCNet method over 24 state-of-the-art models, through extensive quantitative and qualitative comparisons in three widely-used ORSIs-SOD datasets. The source code is available at: \href{https://github.com/CSYSI/UDCNet}{\color{blue} https://github.com/CSYSI/UDCNet}.			
\end{abstract}
\IEEEpeerreviewmaketitle

\section{INTRODUCTION}
\IEEEPARstart{S}{alient} object detection (SOD) mimics the human visual perception mechanism by aiming to identify the most attractive objects or regions within a visual scene from the input data. SOD employed as a pre-processing stage, it has found extensive application in various computer vision tasks, including image segmentation \cite{IS1}, change detection \cite{CD1}, visual tracking \cite{VOT1}, and among others. At the beginning, SOD tasks are focused on natural scene images (NSIs). Over an extended period, some NSIs-SOD methods \cite{PoolNet,CPD,LDF,DCENet,ICON,MINet,RCNet} have achieved tremendous performance. Recently, as remote sensing imaging devices matured, a significant amount of optical remote sensing images (ORSIs) is collected. To better analyze ORSIs, researchers begin to apply SOD tasks to detect remote sensing. ORSIs are more challenging compared to NSIs because they are acquired from satellite or aerial sensors. The objects in ORSIs, as well as their types, scales, illuminations, imaging orientations, and backgrounds, are complex and variable. 

Early ORSIs-SOD methods \cite{T2,T3,T4} usually focused on hand-crafted features or low-level priors ($e.g.$, vision and knowledge-oriented saliency \cite{T2}, color prior \cite{T3}) to detect salient objects from ORSIs. However, the non learnability of traditional ORSIs-SOD methods \cite{T2,T3,T4} results in unsatisfactory results. Subsequently, with the development of neural network and the open source of some large-scale datasets ($i.e.$, ORSSD \cite{ORSSD}, EORSSD \cite{EORSSD}, and ORSI-4199 \cite{ORSI-4199}), numerous ORSIs-SOD methods based on convolutional neural network (CNN) begin to emerge. In particular, Li $et$ $al.$ \cite{ORSSD} first introduced an end-to-end deep network in ORSIs. Later, Zhang and Cong $el$ $al.$ \cite{EORSSD} proposed an end-to-end dense attention fluid network to achieve SOD. Similarly, EMFINet \cite{EMFINet} and MFENet \cite{MFENet} enhanced the discrimination of initial features by conducting multi-scale feature enhancement. In addition, ERPNet \cite{ERPNet} and SEINet \cite{SEINet} utilized edge cues to improve the performance of saliency maps. Furthermore, other CNN-based methods \cite{ACCorNet,CoorNet,SeaNet,HFANet} have reached respectable performance by increasing the difference between objects and backgrounds using different strategies ($e.g.$, semantic guidance \cite{AESINet}, distilling knowledge \cite{SRAL}) in the spatial domain. 

\begin{figure}[t]
	\centering\includegraphics[width=0.48\textwidth,height=3.5cm]{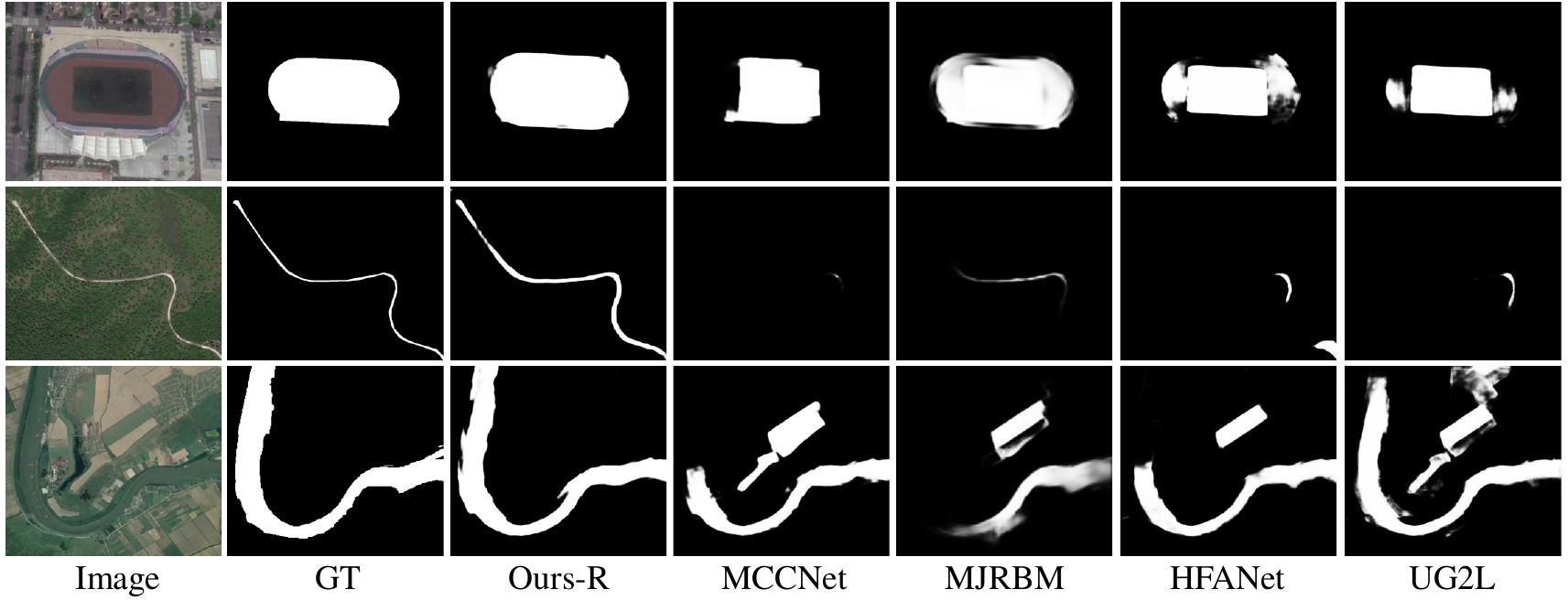}
	\captionsetup{font={small}}
	\caption{ Visual results between the proposed UDCNet model and existing spatial domain-based ORSIs-SOD methods ($i.e.$, MCCNet \cite{MCCNet}, MJRBM \cite{ORSI-4199}, HFANet \cite{HFANet}, and UG2L \cite{UG2L}). }
    \label{Fig.1}
 
\end{figure}

While these ORSIs-SOD methods \cite{ORSSD,EORSSD,ORSI-4199,EMFINet,MCCNet,HFANet} have been successful, we have identified a limitation in their focus primarily on local spatial features within the spatial domain. Spatial features, being local in nature and pixel-based, restrict the pixel's relation solely to itself and its neighboring pixels, resulting in a limited receptive field. Even when certain strategies ($e.g.$, atrous convolution \cite{ASPP}) are adopted to increase the receptive field to aggregate features over a larger region, spatial features are still fundamentally based on local pixel information. This is not conducive to the accurate inference and detection of remote sensing objects (As shown in MCCNet \cite{MCCNet} and MJRBM \cite{ORSI-4199} in Fig. \ref{Fig.1}). Meanwhile, some ORSIs-SOD methods attempt to explore global relationships through utilizing self-attention or different pooling operations within local spatial features, such as HFANet \cite{HFANet}, GeleNet \cite{GeleNet}, GLGCNet\cite{G1}, and UG2L \cite{UG2L}. Specifically, Wang $et$ $al.$ \cite{HFANet} designed a hybrid encoder by combining CNN and Transformer structures to obtain global and local contexts for reducing the disturbance of complex backgrounds. Liu $et$ $al.$ \cite{UG2L} developed a global context block to capture diverse context while modeling long-distance relations for inferring salient objects from ORSIs. Although self-attention mechanisms and various pooling operations allow the model to learn certain global relationships between pixels, local correlations often remain the primary focus when dealing with spatial features. This is largely due to the fact that, in most natural and remote sensing images, the correlations between neighboring pixels are typically more prominent. As depicted in HFANet \cite{HFANet} and UG2L \cite{UG2L} from Fig. \ref{Fig.1}, their performance is still constrained by spatial features with local attributes, leading to incorrect salient objects. How to resolve the locality of features from the spatial domain is significantly meaningful for accurate ORSIs-SOD tasks.

To break the limitations of local spatial features, we introduce an innovative thinking by leveraging the Fourier transform to project these local spatial features into the frequency domain, thereby acquiring frequency features endowed with a global perspective. Extensive researches \cite{FFC,FSI,FSEL} have demonstrated the notion that frequency features derived through the Fourier transform with image-level receptive field inherently possess global properties. Following this, both local spatial and global frequency features undergo a complementary optimization process, culminating in a high-quality representation enriched with both global and local semantics for better inferring remote sensing objects (as indicated in ``Ours-R'' from Fig. \ref{Fig.1}). Specifically, we propose a novel United Domain Cognition Network (UDCNet) for accurate ORSIs-SOD task. In particularly, we design the frequency-spatial domain transformer (FSDT) block that includes four important components, that is, spatial perception self-attention (SPSA), frequency perception self-attention (FPSA), adaptive fusion strategy (AFS), and cross-domain feed-forward network (CDFFN). Technically, the SPSA and FPSA respectively act on local spatial features and global frequency features, correcting the importance information within features through a self-attention structure. Then, we introduce the AFS to adaptively aggregate local spatial features and global frequency features. Next, these global-local information in the proposed CDFFN interact optimization to improve feature expression ability. In addition, we construct the dense semantic excavation (DSE) module to capture into higher-level semantic information by leveraging well-designed atrous convolutions, and utilized dense connections to enhance the correlation, thereby guiding the localization of salient objects. Finally, we design the dual-branch joint optimization (DJO) decoder, which consists of two branches: the edge branch and the saliency branch. The saliency branch enhances the structural information of salient objects through reverse optimization in both the frequency and spatial domains. The edge branch increases the boundary information of objects by introducing an edge refinement. Extensive experimental results on three widely-used benchmarks ($i.e.,$ ORSSD \cite{ORSSD}, EORSSD \cite{EORSSD}, and ORSI-4199 \cite{ORSI-4199}) demonstrate that the proposed UDCNet method outperforms 24 state-of-the-art (SOTA) SOD methods.

Our main contributions are summarized as follows:
\begin{itemize}
	\item We propose a novel United Domain Cognition Network (UDCNet) that generates high-quality representations by integrating global-local information from spatial and frequency domains for better detecting salient objects. 
	
	\item We develop the frequency-spatial domain transformer (FSDT) block, which effectively optimizes and aggregates spatial and frequency features through the collaborative working of SPSA, FPSA, AFS, and CDFFN.
 
	\item We construct the dense semantic excavation (DSE) module, which extracts higher-level semantic to guide the positioning of remote sensing objects by adopting well-designed convolutions and dense connections. 
 
    \item We design the dual-branch joint optimization (DJO) decoder, which utilizes saliency and edge branches to obtain object structure and supplement boundary information for predicting accurate saliency maps.
\end{itemize}

The remainder of this paper is organized as follows. Section II provides the related work, while Section III gives the details of our UDCNet. Experimental results are shown in Section IV. Finally, the concluding remarks are depicted in Section V.
 
\begin{figure*}[t]    
 \centering\includegraphics[width=0.94\textwidth,height=7.3cm]{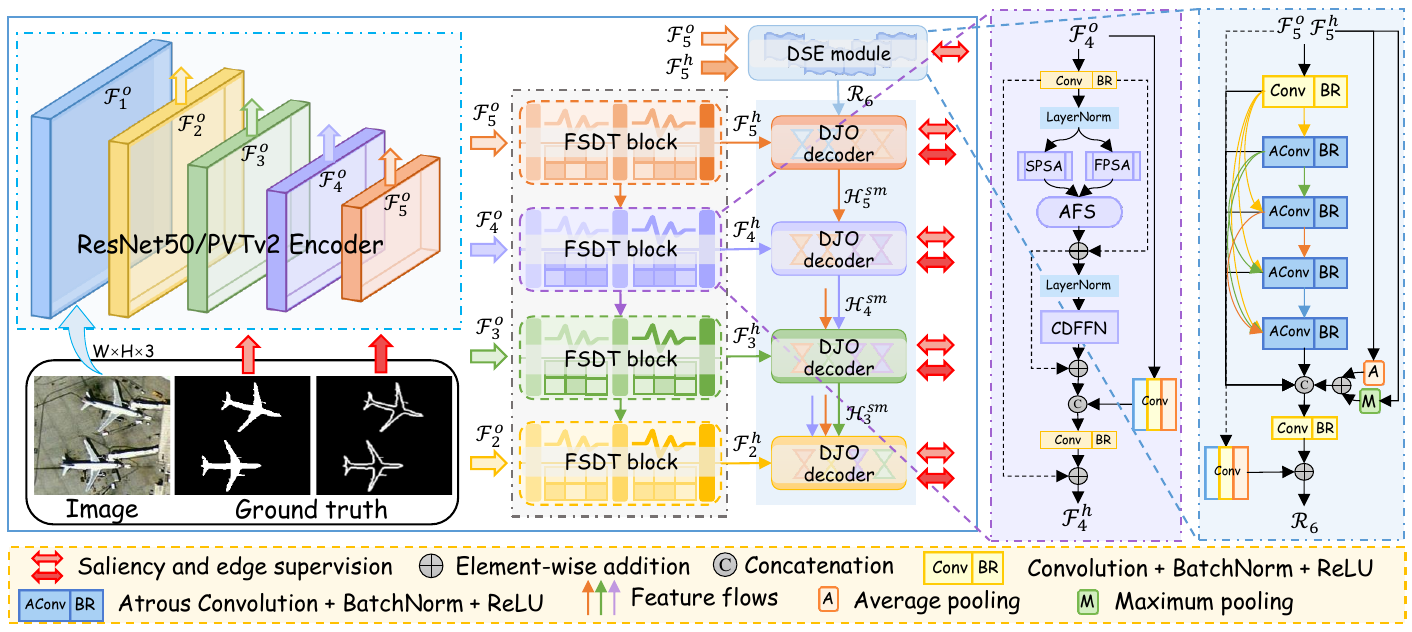}
	\captionsetup{font={small}}
	\caption{Overall framework of our UDCNet method. We use the ResNet50 \cite{ResNet} or PVTv2 \cite{Pvt2} as the backbone, and design the frequency-spatial domain transformer (FSDT) block that contains a spatial perception self-attention (SPSA), a frequency perception self-attention (FPSA), an adaptive fusion strategy (AFS) and a cross-domain feed-forward network (CDFFN) to simultaneously model global relationships and local details. Furthermore, we propose the dense semantic excavation (DSE) module to perform semantic enhancement. In addition, we design the dual-branch joint optimization (DJO) decoder to integrate multi-level features for predicting high-quality saliency maps.}
	\label{Fig.2}
\end{figure*} 

\section{Related Work}

\subsection{Salient Object Detection in ORSIs}
Initially, salient object detection (SOD) is applied in natural scene images (NSIs), and early NSIs-SOD methods \cite{TSOD1,TSOD2} are mainly based on hand-crafted priors. With the rapid development of deep learning, numerous CNN-based NSIs-SOD models \cite{MINet,ITSD,GateNet,ICON,DSP} have been proposed and have achieved extraordinary performance. Later, a large number of remote sensing images have been obtained, leading to widespread attention on SOD methods in optical remote sensing images (ORSIs). Unlike natural images, ORSIs show more challenges ($e.g.$, extreme-scale variation, complex background, and imaging interference). To address these challenges, some ORSIs-SOD models \cite{EMFINet,ERPNet,CoorNet,AESINet,MCCNet} and public datasets have been proposed. In particularly, Li $et$ $al.$ \cite{ORSSD} first provided the public dataset (ORSSD) and designed a encoder-decoder based model for detecting remote sensing objects. Afterward, Zhang $et$ $al.$ \cite{EORSSD} and Tu $et$ $al.$ \cite{ORSI-4199} respectively constructed the EORSSD and ORSI-4199 datasets. Wang $et$ $al.$ \cite{HFANet} designed a adjacent feature aligned module to alleviates the spatial misalignment issue between adjacent features at various scales in ORSIs. Li $et$ $al.$ \cite{MCCNet} embed a multi-content complementation module in an encoder-decoder network to explore the complementarity of multiple content in features for salient objects perception. Gu $et$ $al.$ \cite{sg1} exploited the semantic-guided fusion module in a full-stage manner to make full use of the high-level semantic features and the low-level detailed information for predicting saliency maps. Recently, Li $et$ $al.$ \cite{GeleNet} proposed a transformer-based ORSIs-SOD solution with the global-to-local paradigm to generate more discriminative features. And Bai $et$ $al.$ \cite{G1} developed a global-local-global scheme to stimulated the synergy of global-context-aware and local-context-aware modeling for better predicting salient objects. Although the above methods have achieved impressive performance, they still cannot fully overcome the inherent locality of pixel features in the spatial domain, which makes it difficult to reach the global understanding of the image and the accurate segmentation of remote sensing targets. For that, we break through the limitations of pixel features in the spatial domain by introducing the Fourier transform to map them into frequency domain to obtain global attributes. Then, we conduct a comprehensive analysis of the advantages of both frequency and spatial features and leverage their complementary strengths for better inference of objects.

\subsection{Vision Transformers}
Transformer is an important model for sequence modeling, initially proposed for machine translation tasks \cite{T}. It consists of multiple repetitive modules, each of which contains a self-attention mechanism and a feed-forward network. Later, Transformer models, with their self-attention mechanism that enables simultaneous consideration of all position relationships in the input sequence, have been extensively applied in computer vision tasks, such as image classification \cite{Pvt2,Swin}, camouflaged object detection \cite{FSPNet,GLCONet}, and among others. Specifically, Yuan $et$ $al.$ \cite{ViT} proposed a tokens-to-token vision Transformer to model long-range dependencies from the image patches for performing the image classification. Liu and Lin $et$ $al.$ \cite{Swin} designed a Hierarchical Transformer whose representation is computed with shifted windows to address huge difference between texts and pixels. and Wang $et$ $al.$\cite{Pvt2} constructed pyramid vision Transformer which is first pure Transformer backbone designed for diverse pixel-level dense prediction tasks. Similarly, other Transformer-based methods have been successful in the field of computer vision, $e.g$, OneFormer \cite{One}, Restormer \cite{Restormer}, and so on. In this paper, we introduce the self-attention structure to obtain the correlations between diverse frequency bands and pixels, respectively, and design a cross-domain feed-forward network to learn complex linear relationships in both frequency and spatial domains.

\subsection{Frequency Perceptions}
Recently, image progress in the Fourier space of frequency domain has attracted improving attention, which is capable of obtaining global frequency features effectively due to the periodicity of the Fourier transform for better understanding of image contents. For example, Wang $et$ $al.$ \cite{FP} introduced a method to identify frequency shortcuts, based on culling frequencies that contribute less to image classification. Huang $et$ $al.$ \cite{DF} constructed a new perspective for exposure correction by restoring the representation of different components in the frequency domain. Liu $et$ $al.$ \cite{FSI} proposed a frequency and spatial interactive learning network to introduce frequency learning into under-display camera image restoration. And Wang $et$ $al.$ \cite{DGL} developed a spatial-frequency dynamic graph network to utilize spatial-frequency features to promote forgery detection. In this paper, we utilize the Fourier transform to convert local pixel features into global frequency features in ORSIs-SOD tasks and deeply analyze the correlations among different frequency bands. Then, we adaptively integrate the local spatial and global frequency features to generate powerful representations.

\begin{figure*}[t]
	\centering\includegraphics[width=0.85\textwidth,height=7cm]{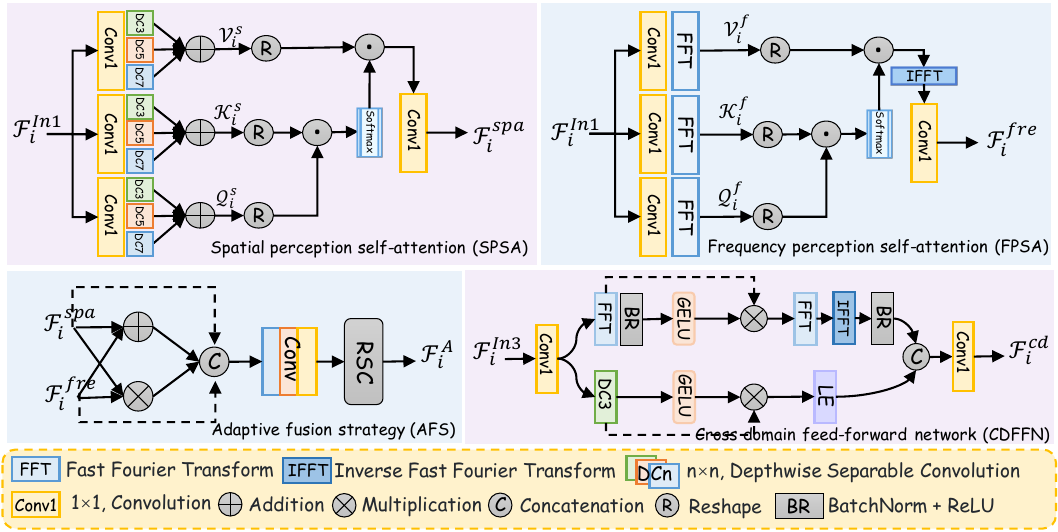}
	\captionsetup{font={small}, justification=raggedright}
	\caption{Details of the frequency-spatial domain transformer block.}
	\label{Fig.3}
\end{figure*}
\section{Proposed UDCNet Method}
We propose a novel UDCNet method to increase the difference between remote sensing objects and backgrounds from the frequency and spatial domains. Specifically, we first describe an overview of our UDCNet in Section A. Then we introduce the details of the three important components ($i.e.$, frequency-spatial domain transformer block, dense semantic excavation module, and dual-branch joint optimization decoder) in Section B-D. Finally, in Section E, we provide the training loss function in the designed UDCNet model.
\subsection{Overview}
Fig. \ref{Fig.2} shows the overall architecture of the proposed UDCNet method, which consists of initial encoder, FSDT block, DSE module, and DJO decoder. Technically, the image $I_c\in\mathbb{R}^{H\times W\times3}$ is input into the initial encoder ($i.e.$, ResNet50 \cite{ResNet} or PVTv2 \cite{Pvt2}) to extract initial multi-level features $\{\mathcal{F} _{i}^{o}\}_{i=1}^{5}$ with the resolution of $W/{2^{i}} \times H/{2^{i}}$. Considering that low-level features contain a large amount of background noise and the larger resolution can result in slower model training, we discard the feature $\mathcal{F} _{1}^{o}$ in the proposed UDCNet method and only use the last four layer features $\{\mathcal{F} _{i}^{o}\}_{i=2}^{5}$. Subsequently, initial features\{$\mathcal{F} _{2}^{o}$, $\mathcal{F} _{3}^{o}$, $\mathcal{F} _{4}^{o}$, $\mathcal{F} _{5}^{o}$\} are sequentially fed into the FSDT block to generate features $\{\mathcal{F} _{i}^{h}\}_{i=2}^{5} $ with abundant global-local knowledge from frequency and spatial domains. Furthermore, we use the DSE module to act on features $\mathcal{F} _{5}^{o}$ and $\mathcal{F} _{5}^{h}$ to capture a higher-level semantic representation $\mathcal{R}_6$ for assisting in locating remote sensing objects. Finally, the DJO decoder is utilized to adaptively integrate diverse information from different layers to gain the powerful edge maps $\{\mathcal{H} _{i}^{em}\}_{i=2}^{5}$ and salient maps $\{\mathcal{H} _{i}^{sm}\}_{i=2}^{5}$ for accurate ORSIs-SOD tasks, and we adopt the binary cross-entropy (BCE) loss, the intersection-over-union (IoU) loss, and edge structure loss to supervise our UDCNet method training.  

\subsection{Frequency-Spatial Domain Transformer Block}
Spatial domain-based methods \cite{ORSSD,EORSSD,EMFINet,ORSI-4199} have achieved excellent performance in ORSIs-SOD tasks, which adopt different optimization strategies to gradually increase the confidence of object pixels and weaken background pixels. However, pixels in the spatial domain often exist within a local perspective, as they only focus on the pixel itself and its immediate information, leading to a partial understanding of the image content and consequently causing sub-optimal results. For that, the FSDT block is proposed. Technically, \textbf{1)} \textit{We transform pixel features of local perspective into frequency features with a Fast Fourier Transform (FFT), enabling a global property.} \textbf{2)} \textit{We correct the importance distribution within the frequency and spatial features through a self-attention structure. } \textbf{3)} \textit{We further enhance the expression ability of feature through joint optimization of frequency and spatial features.} As illustrated in Fig. \ref{Fig.3}, each FSDT block comprises four essential components: spatial perception self-attention (SPSA), frequency perception self-attention (FPSA), adaptive fusion strategy (AFS), and cross-domain feed-forward network (CDFFN).

\subsubsection{Spatial perception self-attention} The purpose of the SPSA is to leverage the self-attention mechanism to gain the relationships between all pixels in the spatial domain. Unlike \cite{Restormer}, SPSA contains more multi-scale information for contextual understanding. Technically, for the input features $\mathcal{F} _{i}^{o}$ and $\mathcal{F} _{i+1}^{h}$($i+1\le 5$), it is first reduced to 128 channels through a 1$\times$1 convolution ($\mathcal{C}_1$). Then, layer normalization ($\mathbb{LN}(\cdot)$) is applied to standardize the data, resulting in the feature $\mathcal{F}_{i}^{In1}$, $i.e.$, $\mathcal{F}_{i}^{In1}=\mathbb{LN}(\mathcal{C}_1\mathbb{C}at(\mathcal{F}_{i}^{o},\mathcal{F}_{i+1}^{h}))$, where $\mathbb{C}at (\cdot, \cdot)$ denotes the concatenation operation. Subsequently, we encode the positions using point-wise convolutions and utilize depth-wise separable convolutions with different kernels ($i.e.$, 3, 5, and 7) to obtain the $query$ ($\mathcal{Q}_{i}^{s}$), $key$ ($\mathcal{K}_{i}^{s}$), and $value$ ($\mathcal{V}_{i}^{s}$) required for the self-attention, which can be formulated as: 
\begin{equation}
	\begin{split}
		&\mathcal{Z}_{i}^{s}=\mathcal{DC}_3^{z}\mathcal{C}_1^{z}\mathcal{F}_{i}^{In1}\oplus\mathcal{DC}_5^{z}\mathcal{C}_1^{z}\mathcal{F}_{i}^{In1}\oplus\mathcal{DC}_7^{z}\mathcal{C}_1^{z}\mathcal{F}_{i}^{In1}, \\
        & \mathcal{Z} = \mathcal{Q},\thinspace\mathcal{K},\thinspace\mathcal{V}. \quad i= 2,3,4,5,
	\end{split}
\end{equation}
where $\mathcal{DC}_n^{z}$ denotes the depthwise separable convolution with $n \times n$ kernel, $\mathcal{C}_1^{z}$ is the point-wise convolution with $1\times1$ kernel, and $\oplus$ presents the element-wise addition operation. And then, we reconstruct the shape of $query$ ($\overline{\mathcal{Q}} _{i}^{s}\in\mathbb{R}^{C\times HW}$) and $key$ ($\overline{\mathcal{K}} _{i}^{s}\in\mathbb{R}^{HW\times C}$) and generate a spatial-attention map $\mathcal{A} _{i}^{s}$ ($\mathcal{A} _{i}^{s} = Softmax(\overline{\mathcal{Q}} _{i}^{s}\odot \overline{\mathcal{K}} _{i}^{s})$ ) using dop-product operation. Finally, the relationships between all pixels in the reshaped value ($\overline{\mathcal{V}} _{i}^{s}\in\mathbb{R}^{C\times HW}$) are modeled based on the attention map $\mathcal{A} _{i}^{s}$ to generate the spatial domain feature $\mathcal{F} _{i}^{spa}$ ($\mathcal{F} _{i}^{spa}=\mathcal{C}_1(\mathcal{A} _{i}^{s}\odot\overline{\mathcal{V}} _{i}^{s})$). $\odot$ denotes the dop-product operation.

\subsubsection{Frequency perception self-attention}
Fast Fourier Transform (FFT) adopts sine and cosine periodic functions as basis functions, enabling the generation of frequency domain features with global characteristics \cite{FFC,FSI}. Therefore, we utilize the FFT to project pixel features from the spatial domain to the frequency domain and use the global spectral features to infer remote sensing objects in the scene. Specifically, taking the optimized feature $\mathcal{F}_{i}^{In1}$ as input, we employ point-wise convolution ($\mathcal{C}_{1}$) and Fast Fourier Transform to obtain frequency domain-based the query ($\mathcal{Q}_{i}^{f}$), key ($\mathcal{K}_{i}^{f}$), and value ($\mathcal{V}_{i}^{f}$). It is important to note that $\mathcal{Q}_{i}^{f}$, $\mathcal{K}_{i}^{f}$, and $\mathcal{V}_{i}^{f}$ are the type of complex numbers, consisting of both real and imaginary components, which are given as:
\begin{equation}
	\begin{split}
		& \mathcal{Z}_{i}^{f}=\overrightarrow{\Phi}|\mathcal{C}_1^{z}\mathcal{F}_{i}^{In1}|,\thinspace \mathcal{Z} = \mathcal{Q},\thinspace\mathcal{K},\thinspace\mathcal{V}. \thinspace\thinspace i= 2,3,4,5,\\
	\end{split}   
\end{equation}
where $\overrightarrow{\Phi}|\cdot|$ denotes the Fast Fourier Transform. Similarly, we use the reconstructed $query$ $\overline{\mathcal{Q}} _{i}^{f}$ and $key$ $\overline{\mathcal{K}} _{i}^{f}$ and perform Softmax activation to generate a complex-type attention map $\mathcal{A} _{i}^{f}$. And then we utilize the attention map $\mathcal{A} _{i}^{f}$ to adaptively recalibrate the reshaped $value$ $\overline{\mathcal{V}} _{i}^{f}$ to obtain the frequency domain-based feature $\mathcal{F} _{i}^{fre}$ through a series of operations, which is defined as:
\begin{equation}
	\begin{split}
		& \mathcal{F}_{i}^{fre}=\mathcal{C}_1\Theta(\overleftarrow{\Phi} |\mathcal{A}_{i}^{f}\odot\overline{\mathcal{V}}_{i}^{f}|), \thinspace\thinspace \\
	\end{split}   
\end{equation}
where $\Theta(\cdot)$ is the magnitude of a complex number, and $\overleftarrow{\Phi} |\cdot|$ denotes the Inverse Fast Fourier Transform.

\subsubsection{Adaptive fusion strategy}
Given a local spatial feature $\mathcal{F} _{i}^{spa}$ and a global frequency feature $\mathcal{F} _{i}^{fre}$, we design an adaptive fusion strategy (AFS) to efficiently aggregate frequency-spatial information. Technically, we employ multiple fusion manners ($i.e.$, element-wise addition, element-wise multiplication, and concatenation) to interact with frequency and spatial features. Firstly, features $\mathcal{F} _{i}^{spa}$ and $\mathcal{F} _{i}^{fre}$ are combined through element-wise addition and multiplication, resulting in features $\mathcal{F}_{i}^{\oplus}$ ($\mathcal{F}_{i}^{\oplus}$=$\mathcal{F} _{i}^{spa}$$\oplus$$ \mathcal{F}_{i}^{fre}$) and $\mathcal{F}_{i}^{\otimes}$ ($\mathcal{F}_{i}^{\otimes}$=$\mathcal{F} _{i}^{spa}$$\otimes$$ \mathcal{F} _{i}^{fre}$), where $\otimes$ denotes the element-wise multiplication. Next, we concatenate features $\mathcal{F} _{i}^{spa}$, $\mathcal{F} _{i}^{fre}$, $\mathcal{F} _{i}^{\oplus}$, and $\mathcal{F} _{i}^{\otimes}$ and gradually reduce the number of channels to 128 using a set of convolutional operations. Subsequently, we introduce a residual spatial-channel attention mechanism \cite{CBAM} that adaptively adjusts the weight to increase the acquisition of significant information in the channel and space to generate feature $\mathcal{F}_{i}^{A}$, as follow:
\begin{equation}
	\begin{split}
		& \mathcal{F}_{i}^{A}=\mathbb{RSC}(\mathbb{C}onv(\mathbb{C}at(\mathcal{F} _{i}^{spa},\mathcal{F} _{i}^{fre},\mathcal{F} _{i}^{\oplus},\mathcal{F} _{i}^{\otimes}))), \\
	\end{split}   
\end{equation}
where $\mathbb{RSC}(\cdot)$ presents the residual spatial-channel attention mechanism \cite{CBAM}, $\mathbb{C}onv(\cdot)$ denotes one 1$\times$1 convolution and two 3$\times$3 convolutions.
\begin{figure}[t]
	\centering\includegraphics[width=0.48\textwidth,height=5cm]{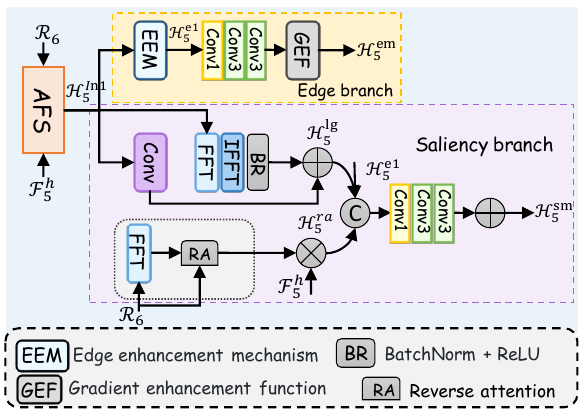}
	\captionsetup{font={small}, justification=raggedright}
	\caption{Details of the dual-branch joint optimization decoder.}
	\label{Fig.4}
\end{figure}
\subsubsection{Cross-domain Feed-Forward Network}
We introduce the CDFFN to enhance the expressive and generalizing capabilities of features within the FSDT block. Our CDFFN structure differs from existing FFN models\cite{Pvt2,Restormer} in that it incorporates both global and local representation reinforcement. Similarly, for the input feature $\mathcal{F}_{i}^{In2}$ ($\mathcal{F}_{i}^{In2}$=$\mathcal{F}_{i}^{A}\oplus\mathcal{C}_1\mathcal{F}_{i}^{o}$), layer normalization is used to normalize the feature, and then we utilize a 1$\times$1 point-wise convolution ($\mathcal{C}_{1}$) to integrate feature flows. Specifically, the fused features $\mathcal{F}_{i}^{In3}$ ($\mathcal{F}_{i}^{In3}$=$\mathcal{C}_1\mathbb{LN}(\mathcal{F}_{i}^{In2})$) are input into two branches. The local branch consists of a series of depthwise separable convolutions and a local enhancement operation for non-linear transformations. In the global branch, we first apply an FFT operation to project pixel features into the frequency domain to focus more on global information and employ a gate mechanism and GELU for non-linear transformations. Subsequently, the FFT is adopted again to transform and encode the frequency features, blending these bands, and thereby aiding the network in extracting deeper global spectral information. Next, the IFFT operation converts the frequency feature back to the original domain. The optimized spatial feature $\mathcal{F}_{i}^{l}$ and frequency feature $\mathcal{F}_{i}^{g}$ are concatenated, followed by a 1$\times$1 point-wise convolution ($\mathcal{C}_{1}$) to generate powerful features $\mathcal{F}_{i}^{cd}$, as follows:
\begin{equation}
	\begin{split}	         &\mathcal{F}_{i}^{cd}=\mathcal{C}_1\mathbb{C}at(\mathcal{F}_{i}^{l},\mathcal{F}_{i}^{g}), i=2,3,4,5 \\ &\mathcal{F}_{i}^{g}=\Theta(\overleftarrow{\Phi}|\overrightarrow{\Phi}|\Psi(\Theta(\overrightarrow{\Phi}|\mathcal{F}_{i}^{In3}|))\otimes\Theta(\overrightarrow{\Phi}|\mathcal{F}_{i}^{In3}|)||),\\
     &\mathcal{F}_{i}^{l}=\mathbb{LE}(\Psi (\mathcal{DC}_3(\mathcal{F}_{i}^{In3}))\otimes\mathcal{DC}_3(\mathcal{F}_{i}^{In3})), \\
	\end{split}   
\end{equation}
where $\Psi(\cdot)$ is GELU non-linear function, $\mathbb{LE}(\cdot)$ represents the local enhancement operation that contains multiple atrous convolutions with different receptive fields. Finally, we introduce a series of residual connection to increase feature diversity to generate feature $\mathcal{F}_{i}^{h}$, that is, $\mathcal{F}_{i}^{h} = \mathcal{C}_1\mathbb{C}at(\mathcal{F}_{i}^{cd} \oplus \mathcal{F}_{i}^{In2},\mathbb{C}onv(\mathcal{F}_{i}^{o})) \oplus \mathcal{C}_1 \mathcal{F}_{i}^{o}$. Through the optimization of the FSDT block, the original feature $\mathcal{F}_{i}^{o}$ increases a large amount of local and global perception information, which is beneficial for assisting in segmentation of remote sensing targets.
\begin{table*}[t]
\renewcommand{\arraystretch}{1}
	\setlength{\tabcolsep}{4pt}
	\centering
	\caption{ Quantitative results on three ORSIs-SOD benchmarks. The best three results are shown in {\color{red} red}, {\color{green} green}, and {\color{blue} blue}. ``$\uparrow$/$\downarrow$'' present that a higher/lower value is better. ``Ours-R'' and  ``Ours-P'' denote ResNet50 \cite{ResNet} or PVT \cite{Pvt2} as backbone.}
	\resizebox*{0.90\textwidth}{75mm}{
\begin{tabular}{cccccccccccccccccccc}
\hline\hline
\multicolumn{1}{c|}{\multirow{2}{*}{\textbf{Methods}}} & \multicolumn{1}{c|}{\multirow{2}{*}{\textbf{Publication}}} & \multicolumn{6}{c|}{\textbf{ORSSD (200 images)}}                                   & \multicolumn{6}{c|}{\textbf{EORSSD (600 images)}}                                  & \multicolumn{6}{c}{\textbf{ORSI-4199 (2199 images)}}           \\
\multicolumn{1}{c|}{}                        & \multicolumn{1}{c|}{}                             & \cellcolor{blue!15}$\mathcal{M}$$\downarrow$   & \cellcolor{blue!15}$F_{\varphi}^{m}$$\uparrow$   & \cellcolor{blue!15}$F_{\varphi}^{a}$$\uparrow$   & \cellcolor{blue!15}$F_{\varphi}^{w}$$\uparrow$   & \cellcolor{blue!15}$S_m$$\uparrow$     & \multicolumn{1}{c|}{\cellcolor{blue!15}$E_m$$\uparrow$}     & \cellcolor{blue!15}$\mathcal{M}$$\downarrow$   & \cellcolor{blue!15}$F_{\varphi}^{m}$$\uparrow$   & \cellcolor{blue!15}$F_{\varphi}^{a}$$\uparrow$   & \cellcolor{blue!15}$F_{\varphi}^{w}$$\uparrow$   & \cellcolor{blue!15}$S_m$$\uparrow$     & \multicolumn{1}{c|}{\cellcolor{blue!15}$E_m$$\uparrow$}     & \cellcolor{blue!15}$\mathcal{M}$$\downarrow$   & \cellcolor{blue!15}$F_{\varphi}^{m}$$\uparrow$   & \cellcolor{blue!15}$F_{\varphi}^{a}$$\uparrow$   & \cellcolor{blue!15}$F_{\varphi}^{w}$$\uparrow$   & \cellcolor{blue!15}$S_m$$\uparrow$     & \multicolumn{1}{c}{\cellcolor{blue!15}$E_m$$\uparrow$}     \\ \hline
\multicolumn{20}{c}{Natural Scene Images-SOD Methods}                                                                                                                                                                                                                                                                                     \\ \hline
\multicolumn{1}{c|}{MINet \cite{MINet}}                   & \multicolumn{1}{c|}{2020,CVPR}                    & 0.0144 & 0.8927 & 0.8251 & 0.8443 & 0.8945 & \multicolumn{1}{c|}{0.9305} & 0.0093 & 0.8583 & 0.7705 & 0.8132 & 0.8690 & \multicolumn{1}{c|}{0.9007} & -      & -      & -      & -      & -      & -      \\ 
\multicolumn{1}{c|}{ITSD \cite{ITSD}}                    & \multicolumn{1}{c|}{2020,CVPR}                    & 0.0165 & 0.8847 & 0.8068 & 0.8381 & 0.8948 & \multicolumn{1}{c|}{0.9269} & 0.0106 & 0.8690 & 0.7421 & 0.8114 & 0.8672 & \multicolumn{1}{c|}{0.8996} & -      & -      & -      & -      & -      & -      \\ 
\multicolumn{1}{c|}{CSNet \cite{CSF}}                   & \multicolumn{1}{c|}{2020,ECCV}                    & 0.0186 & 0.8923 & 0.7615 & 0.7243 & 0.8894 & \multicolumn{1}{c|}{0.9069} & 0.0169 & 0.8486 & 0.6319 & 0.5796 & 0.8331 & \multicolumn{1}{c|}{0.8347} & 0.0524 & 0.8404 & 0.7162 & 0.6861 & 0.8227 & 0.8446 \\ 
\multicolumn{1}{c|}{GateNet \cite{GateNet}}                 & \multicolumn{1}{c|}{2020,ECCV}                    & 0.0137 & 0.8972  & 0.8229 & 0.8567 & 0.9102 & \multicolumn{1}{c|}{0.9313} & 0.0095 & 0.8716 & 0.7109 & 0.8127 & 0.8733 & \multicolumn{1}{c|}{0.8753} & -      & -      & -      & -      & -      & -      \\ 
\multicolumn{1}{c|}{PA-KRN \cite{PAKRN}}                  & \multicolumn{1}{c|}{2021,AAAI}                    & 0.0139 & 0.8956 & 0.8548 & 0.8678 & 0.9145 & \multicolumn{1}{c|}{0.9411} & 0.0104 & 0.8750 & 0.7993 & 0.8391 & 0.8802 & \multicolumn{1}{c|}{0.9269} & 0.0382 & 0.8592 & 0.8200 & 0.8107 & 0.8427 & 0.9165 \\ 
\multicolumn{1}{c|}{SAMNet \cite{SAMNet}}                  & \multicolumn{1}{c|}{2021,TIP}                     & 0.0132 & 0.8062 & 0.6114 & 0.7010 & 0.8304 & \multicolumn{1}{c|}{0.8111} & 0.0217 & 0.8335 & 0.6843 & 0.7297 & 0.8680 & \multicolumn{1}{c|}{0.8556} & 0.0432 & 0.8435 & 0.7744 & 0.7732 & 0.8353 & 0.8940 \\ 
\multicolumn{1}{c|}{VST \cite{VST}}                     & \multicolumn{1}{c|}{2021, ICCV}                   & 0.0094 & 0.9171 & 0.8262 & 0.8721 & 0.9267 & \multicolumn{1}{c|}{0.9402} & 0.0067 & 0.8812 & 0.7089 & 0.8188 & 0.8829 & \multicolumn{1}{c|}{0.8789} & \color{blue}0.0281 & 0.8829 & 0.7947 & 0.8353 & 0.8733 & 0.9067 \\ 
\multicolumn{1}{c|}{DPORTNet \cite{DPORTNet}}                & \multicolumn{1}{c|}{2022,TIP}                     & 0.0220 & 0.8597 & 0.7970 & 0.7978 & 0.8754 & \multicolumn{1}{c|}{0.8968} & 0.0150 & 0.8555 & 0.7545 & 0.7906 & 0.8585 & \multicolumn{1}{c|}{0.8859} & 0.0569 & 0.8097 & 0.7554 & 0.7410 & 0.8041 & 0.8655 \\ 
\multicolumn{1}{c|}{DNTD \cite{DNRF}}                    & \multicolumn{1}{c|}{2022,SCIS}                    & 0.0217 & 0.8477 & 0.7645 & 0.7796 & 0.8627 & \multicolumn{1}{c|}{0.8881} & 0.0113 & 0.8448 & 0.7288 & 0.7895 & 0.8625 & \multicolumn{1}{c|}{0.8762} & 0.0425 & 0.8523 & 0.8065 & 0.7917 & 0.8390 & 0.9034 \\ 
\multicolumn{1}{c|}{ICON-P \cite{ICON}}                  & \multicolumn{1}{c|}{2023,TPAMI}                   & 0.0116 & 0.9037 & 0.8444 & 0.8667 & 0.9162 & \multicolumn{1}{c|}{0.9425} & 0.0073 & 0.8705 & 0.8065 & 0.8431 & 0.8821 & \multicolumn{1}{c|}{0.9216} & 0.0282 & \color{blue}0.8870 & 0.8531 & \color{blue}0.8521 & 0.8692 & \color{blue}0.9443 \\ \hline
\multicolumn{20}{c}{Optical Remote Sensing Images-SOD Methods}                                                                                                                                                                                                                                                                                        \\ \hline
\multicolumn{1}{c|}{LVNet \cite{ORSSD}}                   & \multicolumn{1}{c|}{2019,TGRS}                    & 0.0207 & 0.8414 & 0.7506 & 0.7746 & 0.8730 & \multicolumn{1}{c|}{0.9230} & 0.0145 & 0.8051 & 0.6306 & 0.7021 & 0.8355 & \multicolumn{1}{c|}{0.8476} & -      & -      & -      & -      & -      & -      \\ 
\multicolumn{1}{c|}{DAFNet \cite{EORSSD}}                  & \multicolumn{1}{c|}{2021,TIP}                     & 0.0113 & 0.9032 & 0.7876 & 0.8441 & 0.9119 & \multicolumn{1}{c|}{0.9195} & \color{blue}0.0060 & 0.8673 & 0.6423 & 0.7831 & 0.8830 & \multicolumn{1}{c|}{0.8154} & -      & -      & -      & -      & -      & -      \\ 
\multicolumn{1}{c|}{EMFINet \cite{EMFINet}}                 & \multicolumn{1}{c|}{2022,TGRS}                      & 0.0109 & 0.9070 & 0.8617 & 0.8809 & 0.9267 & \multicolumn{1}{c|}{0.9662} & 0.0084 & 0.8822 & 0.7984 & 0.8487 & 0.8891 & \multicolumn{1}{c|}{0.9495} & 0.0330 & 0.8717 & 0.8186 & 0.8301 & 0.8612 & 0.9135 \\ 
\multicolumn{1}{c|}{EPRNet \cite{ERPNet}}                  & \multicolumn{1}{c|}{2022,TCYB}                    & 0.0135 & 0.9040 & 0.8356 & 0.8636 & 0.9153 & \multicolumn{1}{c|}{0.9526} & 0.0089 & 0.8769 & 0.7554 & 0.8246 & 0.8812 & \multicolumn{1}{c|}{0.9225} & 0.0357 & 0.8716 & 0.8024 & 0.8156 & 0.8606 & 0.9039 \\ 
\multicolumn{1}{c|}{HFANet \cite{HFANet}}                  & \multicolumn{1}{c|}{2022,TGRS}                    & 0.0092 & 0.9174 & 0.8819 & 0.8941 & 0.9297 & \multicolumn{1}{c|}{0.9522} & 0.0070 & \color{blue}0.8947 & \color{blue}0.8365 & \color{green}0.8708 & \color{green}0.8957 & \multicolumn{1}{c|}{0.9338} & 0.0314 & 0.8819 & 0.8323 & 0.8453 & 0.8702 & 0.9191 \\ 
\multicolumn{1}{c|}{MJRBM \cite{ORSI-4199}}                   & \multicolumn{1}{c|}{2022,TGRS}                    & 0.0163 & 0.8932 & 0.8022 & 0.8435 & 0.9097 & \multicolumn{1}{c|}{0.9337} & 0.0099 & 0.8765 & 0.7066 & 0.8134 & 0.8786 & \multicolumn{1}{c|}{0.8899} & 0.0374 & 0.8668 & 0.7995 & 0.8055 & 0.8530 & 0.9088 \\ 
\multicolumn{1}{c|}{ACCoNet \cite{ACCorNet}}                 & \multicolumn{1}{c|}{2022,TCYB}                    & 0.0088 & 0.9196 & 0.8806 & 0.8968 & 0.9335 & \multicolumn{1}{c|}{0.9756} & 0.0074 & 0.8927 & 0.7969 & 0.8592 & 0.8898 & \multicolumn{1}{c|}{0.9468} & 0.0314 & 0.8818 & 0.8581 & 0.8446 & \color{blue}0.8711 & 0.9402 \\ 
\multicolumn{1}{c|}{CorrNet \cite{CoorNet}}                 & \multicolumn{1}{c|}{2022, TGRS}                   & 0.0098 & 0.9171 & 0.8875 & 0.8963 & 0.9286 & \multicolumn{1}{c|}{0.9756} & 0.0083 & 0.8907 & 0.8311 & 0.8621 & 0.8888 & \multicolumn{1}{c|}{\color{blue}0.9598} & 0.0366 & 0.8732 & 0.8534 & 0.8283 & 0.8562 & 0.9300 \\ 
\multicolumn{1}{c|}{MCCNet \cite{MCCNet}}                  & \multicolumn{1}{c|}{2022,TGRS}                    & 0.0087 & \color{blue}0.9214 & \color{green}0.8957 & \color{blue}0.9024 & 0.9334 & \multicolumn{1}{c|}{\color{green}0.9770} & 0.0066 & \color{green}0.8991 & 0.8137 & 0.8665 & 0.8954 & \multicolumn{1}{c|}{0.9546} & 0.0316 & 0.8809 & \color{blue}0.8592 & 0.8468 & 0.8682 & 0.9401 \\ 
\multicolumn{1}{c|}{SeaNet \cite{SeaNet}}                  & \multicolumn{1}{c|}{2023,TRGS}                    & 0.0105 & 0.8994 & 0.8625 & 0.8730 & 0.9174 & \multicolumn{1}{c|}{\color{blue}0.9706} & 0.0073 & 0.8743 & 0.8304 & 0.8501 & 0.8838 & \multicolumn{1}{c|}{\color{green}0.9608} & 0.0308 & 0.8784 & 0.8557 & 0.8419 & 0.8661 & 0.9394 \\ 
\multicolumn{1}{c|}{AESINet \cite{AESINet}}                 & \multicolumn{1}{c|}{2023,TRGS}                    & \color{blue}0.0085 & 0.9194 & 0.8640 & 0.8953 & \color{blue}0.9352 & \multicolumn{1}{c|}{0.9538} & 0.0064 & 0.8864 & 0.7909 & 0.8533 & \color{blue}0.8956 & \multicolumn{1}{c|}{0.9283} & -      & -      & -      & -      & -      & -      \\ 
\multicolumn{1}{c|}{SARL \cite{SRAL}}                    & \multicolumn{1}{c|}{2023,TGRS}                    & 0.0107 & 0.9064 & 0.8514 & 0.8744 & 0.9225 & \multicolumn{1}{c|}{0.9442} & 0.0069 & 0.8796 & 0.8146 & 0.8505 & 0.8869 & \multicolumn{1}{c|}{0.9256} & 0.0307 & 0.8795 & 0.8230 & 0.8419 & 0.8678 & 0.9193 \\ 
\multicolumn{1}{c|}{SASOD \cite{SASOD}}                   & \multicolumn{1}{c|}{2023,TGRS}                    & 0.0101 & 0.9128 & 0.8752 & 0.8894 & 0.9240 & \multicolumn{1}{c|}{0.9491} & 0.0065 & 0.8930 & \color{green}0.8371 & \color{blue}0.8699 & 0.8945 & \multicolumn{1}{c|}{0.9308} & 0.0299 & 0.8835 & 0.8399 & 0.8452 & \color{blue}0.8711 & 0.9218 \\ 
\multicolumn{1}{c|}{UG2L \cite{UG2L}}                    & \multicolumn{1}{c|}{2023,GRSL}                    & 0.0098 & 0.9174 & 0.8764 & 0.8917 & 0.9303 & \multicolumn{1}{c|}{0.9510} & 0.0069 & 0.8933 & 0.8350 & 0.8695 & 0.8955 & \multicolumn{1}{c|}{0.9327} & 0.0318 & 0.8767 & 0.8481 & 0.8380 & 0.8668 & 0.9251 \\ \hline
\multicolumn{1}{c|}{\textbf{Ours-R}}                  & \multicolumn{1}{c|}{-}                            & \color{green}0.0068 & \color{green}0.9267 & \color{blue}0.8932 & \color{green}0.9093 & \color{green}0.9389 & \multicolumn{1}{c|}{\color{green}0.9770} & \color{green}0.0056 & 0.8903 & 0.8211 & 0.8615 & \color{green}0.8957 & \multicolumn{1}{c|}{0.9547} &\color{green} 0.0266 & \color{green}0.8899 & \color{green}0.8648 & \color{green}0.8609 & \color{green}0.8744 & \color{green}0.9475 \\ \hline
\multicolumn{1}{c|}{\textbf{Ours-P}}                & \multicolumn{1}{c|}{-}                            & \color{red}{0.0052} & \color{red}0.9344 & \color{red}0.9048 & \color{red}0.9204 & \color{red}0.9460 & \multicolumn{1}{c|}{\color{red}0.9862} & \color{red}0.0050 & \color{red}0.8981 & \color{red}0.8417 & \color{red}0.8778 & \color{red}0.9017 & \multicolumn{1}{c|}{\color{red}0.9658} & \color{red}0.0237 & \color{red}0.8961 & \color{red}0.8729 & \color{red}0.8709 & \color{red}0.8811 & \color{red}0.9547 \\ \hline\hline
\end{tabular} }

\label{table1}

\end{table*}

\begin{figure*}[]
	\scriptsize
	\centering
	\begin{tabular}{ccc}		
		\includegraphics[width=0.23\textwidth,height=2.9cm]{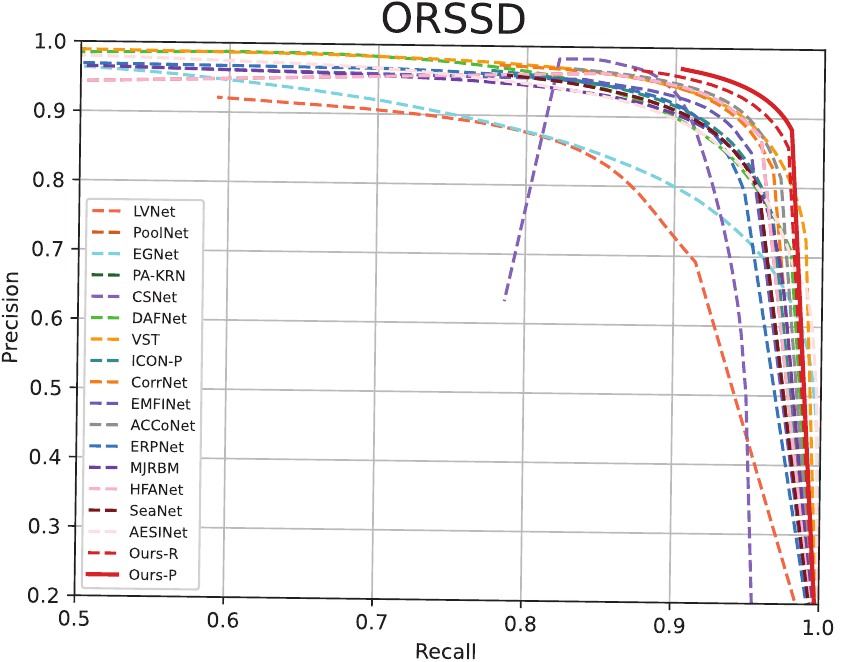}\
		&\includegraphics[width=0.23\textwidth,height=2.9cm]{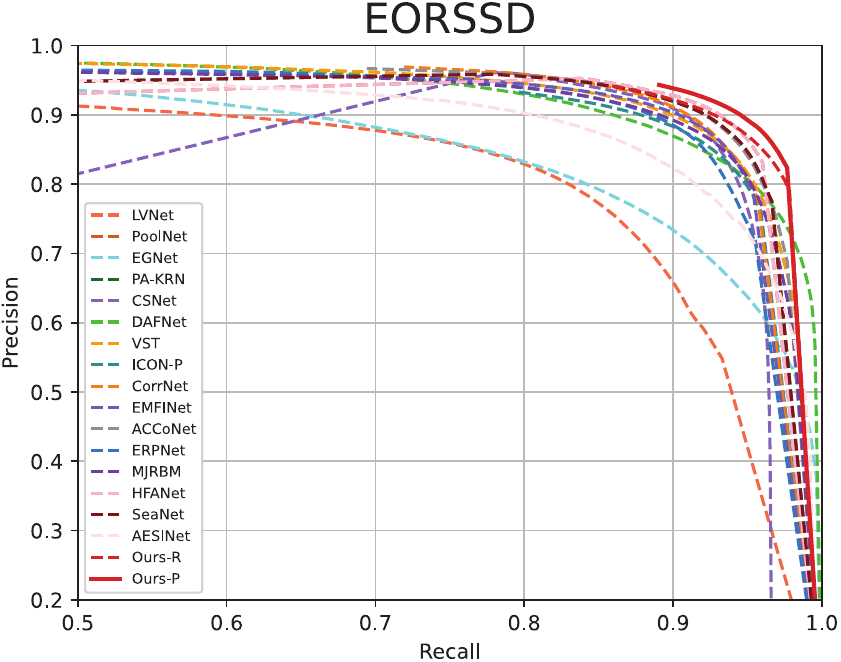}\
		&\includegraphics[width=0.23\textwidth,height=2.9cm]{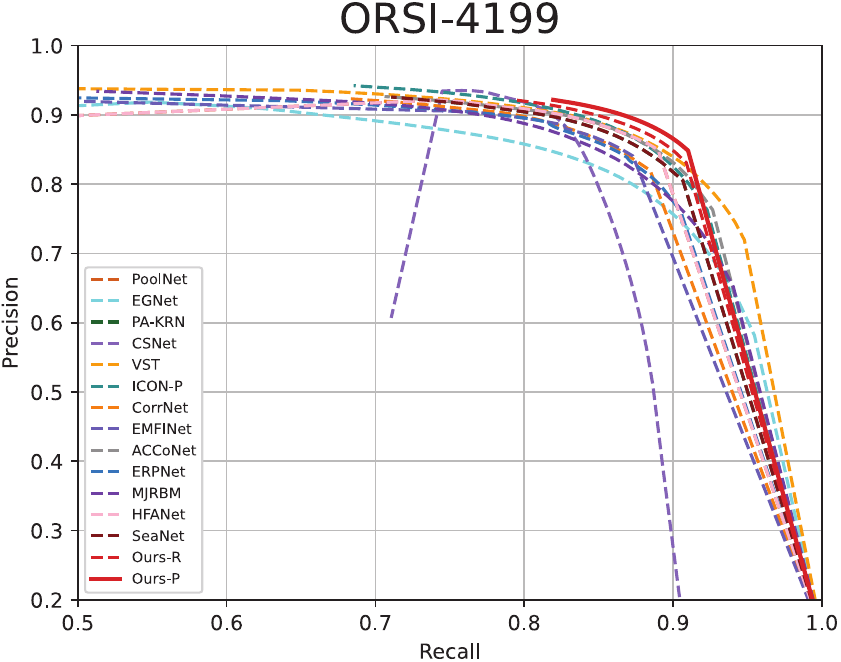}\\	
	\end{tabular}
	\begin{tabular}{ccc}		
		\includegraphics[width=0.23\textwidth,height=2.9cm]{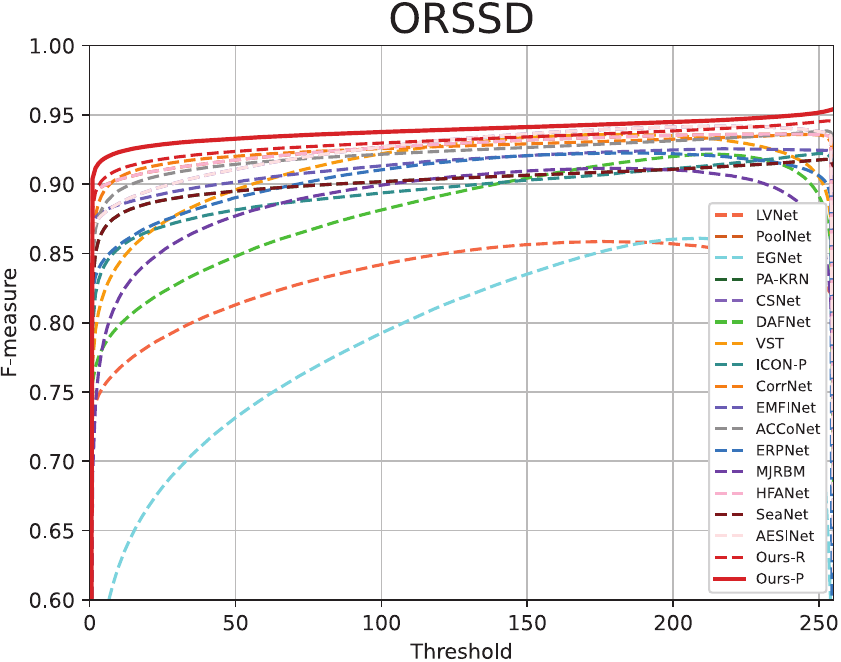}\
		&\includegraphics[width=0.23\textwidth,height=2.9cm]{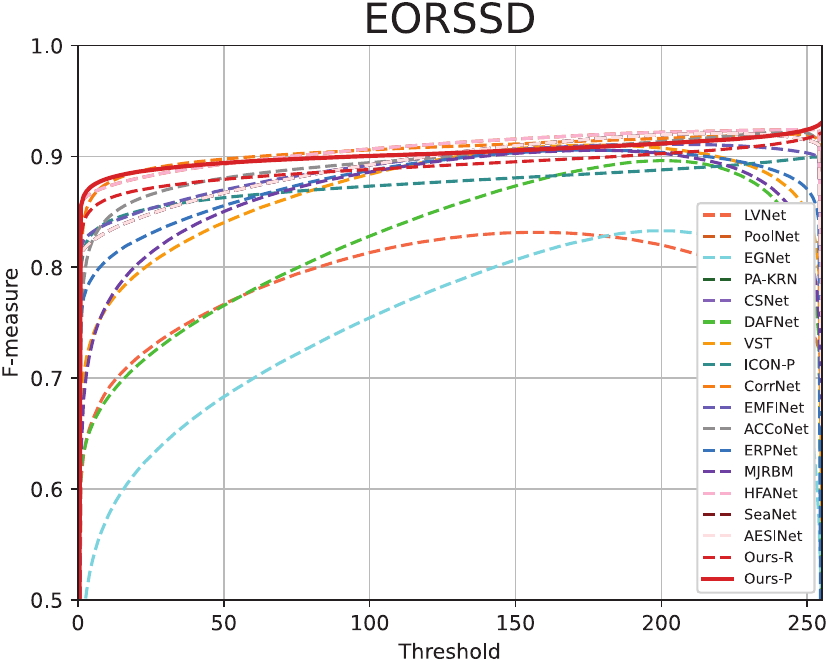}\
		&\includegraphics[width=0.23\textwidth,height=2.9cm]{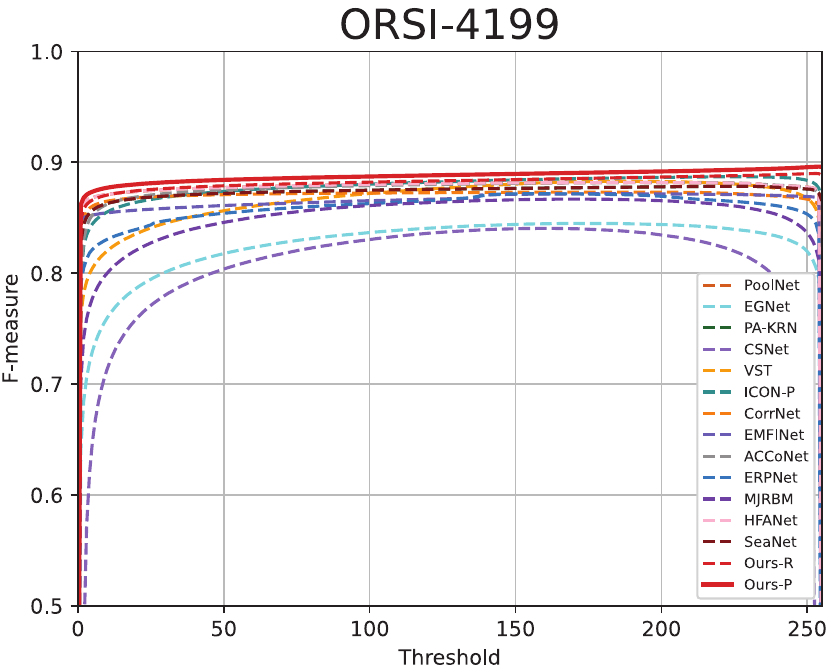}\\
	\end{tabular}		
	\captionsetup{font={small}, justification=raggedright}
	\caption{Quantitative resluts of the $PR$ and $F_m$ curves for UDCNet and other SOTA methods on three ORSIs-SOD datasets.}
	\label{Fig.5}

\end{figure*}
\begin{figure*}[]
	\centering\includegraphics[width=0.84\textwidth,height=7cm]{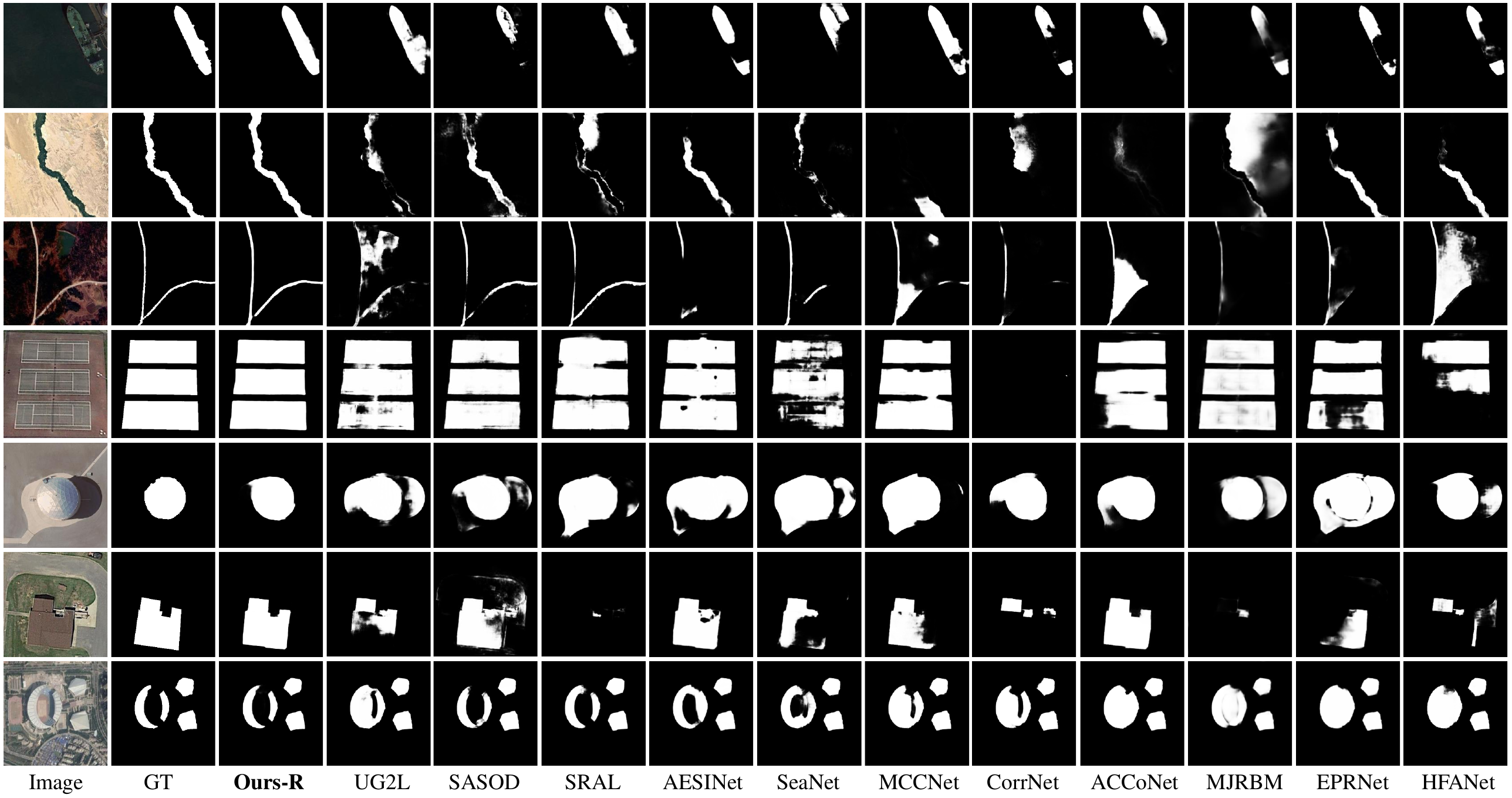}
	\captionsetup{font={small}, justification=raggedright}
	\caption{Qualitative comparisons of the proposed UDCNet method and 11 ORSIs-SOD methods.}
	\label{Fig.6}
\end{figure*}
\begin{figure}[]
    \centering\includegraphics[width=0.5\textwidth,height=3.1cm]{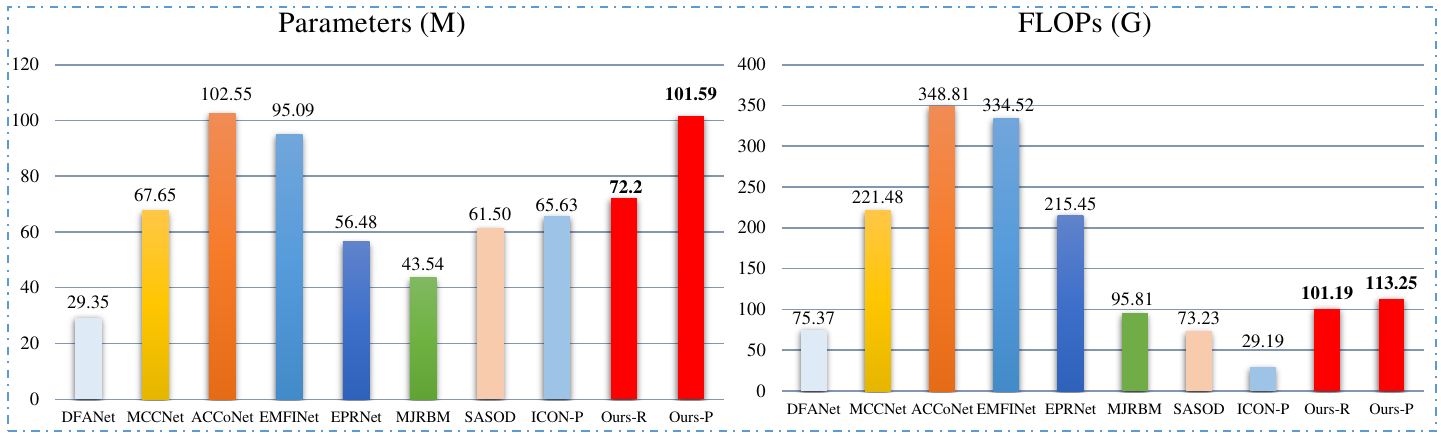}
    \captionsetup{font={small}, justification=raggedright}
    \caption{Efficiency analysis of the proposed UDCNet method and existing ORSIs-SOD methods.}
    \label{Fig.7}
\end{figure}
\subsection{Dense semantic excavation module}
High-level features often include a significant amount of semantic knowledge that helps in identifying remote sensing objects in images. Based on this, we propose the DSE module to extract higher-level semantic information with different scales, which guides subsequent reasoning. Unlike the atrous spatial pyramid pooling (ASPP) \cite{ASPP} and the receptive field block (RFB) \cite{RFB}, we enhance the correlation between semantic information from different scales. Specifically, we utilize four atrous convolutions ($\mathcal{AC}_{3}^{n}$) that set the dilation rate to n ($i.e.$, n = 3, 6, 12, 18) to obtain rich semantic information from high-level features $\mathcal{F}_{5}^{o}$ and $\mathcal{F}_{5}^{h}$. As depicted in the right side of Fig. \ref{Fig.2}, we adopt a dense connection, utilizing the gradual integration of small receptive fields into large receptive fields to strengthen the relationship between multi-scale semantic features. This can enhance the correlation between features and increase the diversity of information to generate feature $\mathcal{R}_{6}^{In1}$ with abundant semantic, which can be defined as: 
\begin{equation}
		\begin{split}  &\mathcal{R}_{6}^{In1}=\mathbb{C}at(\mathcal{R}_{6}^{1},\mathcal{R}_{6}^{3},\mathcal{R}_{6}^{6},\mathcal{R}_{6}^{12},\mathcal{R}_{6}^{18},\mathcal{R}_{6}^{am}), \\
		\end{split}  
\end{equation}
where $\mathcal{R}_{6}^{1}$ is generated by 1$\times$1 point-wise convolution ($\mathcal{C}_{1}$), $\mathcal{R}_{6}^{n}$ is produced by 3$\times$3 atrous convolution ($\mathcal{AC}_{3}^{n}$) with dilation rate $n$ (3, 6, 12, 18), $\mathcal{R}_{6}^{am}$ is obtained by average pooling and maximum pooling. Subsequently, we gradually reduce the channels of feature $\mathcal{R}_{6}^{In1}$ through a group of 3$\times$3 convolutions. Additionally, we introduce a residual connection to increase feature diversity, and ultimately gain a  coarse feature map $\mathcal{R}_{6}$ with 1-channel. It can be formulated as:
\begin{equation}
		\begin{split}  
            &\mathcal{R}_{6}=\mathcal{C}_1^{\times 2}(\mathcal{C}_3^{\times 3}\mathcal{R}_{6}^{In1}\oplus\mathbb{C}onv(\mathbb{C}at(\mathcal{F}_{5}^{o},\mathcal{F}_{5}^{h}))), \\
		\end{split}  
\end{equation}
where $\mathcal{C}_{k}^{\times n}$ denotes that there is $n$ convolution with $k\times k$ kernel.
	
\subsection{Dual-branch Joint Optimization Decoder}
Different levels of features contain complementary information ($i.e.$, visual semantics in high-level features and edge details in low-level features), and high-quality saliency maps need to include complete objects as well as fine-grained edge structures. How to integrate this information is crucial for accurate ORSI-SOD tasks. To this end, we design a dual-branch joint optimization (DJO) decoder which is applied on saliency branch and edge branch to strengthen representations (as shown in Fig. \ref{Fig.4}). A saliency branch is utilized to locate and segment remote sensing objects through a series of optimizations in frequency and spatial domain. And an edge branch is adopted to help refine the boundary in the saliency branch.

Specifically, we use the optimized features $\mathcal{R}_6$ and $\mathcal{F}_{5}^{h}$ as inputs. Firstly, we aggregate the feature $\mathcal{H}_{5}^{In1}$ ($\mathcal{H}_{5}^{In1}$=$\mathbb{AFS}(\mathcal{R}_6,\mathcal{F}_{5}^{h})$) by employing the adaptive fusion strategy (AFS). In the saliency branch, we utilize a set of convolution operations for local optimization. Subsequently, we project local spatial features onto the frequency domain using the Fourier operations ($i.e.$, FFT and IFFT) to enhance global patterns and stable distribution to obtain feature $\mathcal{H}_{5}^{lg}$ with both local and global properties. Its formulation is defined as:
\begin{equation}
		\begin{split}         &\mathcal{H}_{5}^{lg}=\mathbb{C}onv(\mathcal{H}_{5}^{In1})\oplus\Theta(\overleftarrow{\Phi}|\overrightarrow{\Phi}|\mathcal{H}_{5}^{In1}||). \\
		\end{split}  
\end{equation}
Additionally, we design a spatial-frequency reverse attention mechanism that extends the reverse attention mechanism to both frequency and spatial domains. This mechanism obtains a hybrid domain attention map that helps locate remote sensing objects and produces the feature $\mathcal{H}_{5}^{ra}$, which can is given as:
\begin{equation}
		\begin{split}         
        &\mathcal{H}_{5}^{ra}=\mathcal{F}_{5}^{h}\otimes(\mathbb{RA}(\mathcal{R}_{6})\oplus\mathbb{RA}(\Theta(\overrightarrow{\Phi}|\mathcal{R}_{6}|))) , \\
		\end{split}  
\end{equation}
where $\mathbb{RA}(\cdot)$ denote the reverse attention \cite{RASNet} that is defined as (-1 $\otimes$ Sigmoid $(\cdot)$ $\oplus$ 1). For the edge branch, we start by using the edge enhancement mechanism (EEM) that contains small kernel convolutions and the spatial attention to gradually refine its edge details to obtain feature $\mathcal{H}_{5}^{e1}$, as follows:
\begin{equation}
		\begin{split}         
        &\mathcal{H}_{5}^{e1}=\mathcal{C}_{1}\mathbb{C}at(\mathbb{SA}(\mathbb{SC}(\mathcal{H}_{5}^{In1})),\mathbb{C}onv(\mathbb{SC}(\mathcal{H}_{5}^{In1}))) , \\
		\end{split}  
\end{equation}
where $\mathbb{SA}(\cdot)$ denotes the spatial attention \cite{sa}, $\mathbb{SC}(\cdot)$ is the small kernel convolution that contains one $1 \times 1$ convolution, one $3 \times 3 $ convolution, and one $1 \times 1$ convolution. Afterwards, we utilize a set of convolutions with a $3 \times 3$ kernel to generate an edge map and apply a gradient enhancement function (GEF) to enhance the edge structure, obtaining an unactivated edge map $\mathcal{H}_{5}^{em}$ a by:
\begin{equation}
		\begin{split}         
        &\mathcal{H}_{5}^{em}=\mathbb{GEF}(\mathcal{C}_{3}^{\times3}(\mathcal{H}_{5}^{e1})), \\
		\end{split}  
\end{equation}
where $\mathcal{C}_{k}^{\times n}$ presents that there is $n$ convolution with $k\times k$ kernel, $\mathbb{GEF}(\cdot)$ denotes the gradient enhancement function. For the final saliency map $\mathcal{H}_{5}^{sm}$, we first concatenate the optimized features $\mathcal{H}_{5}^{lg}$, $\mathcal{H}_{5}^{ra}$, and $\mathcal{H}_{5}^{e1}$ and use a set of convolutional operations to predict the saliency map and aggregate the high-level saliency map $\mathcal{R}_6$. The process is as follows:
\begin{equation}
		\begin{split}         &\mathcal{H}_{5}^{sm}=\mathcal{C}_{3}^{\times3}\mathbb{C}at(\mathcal{H}_{5}^{lg},\mathcal{H}_{5}^{ra},\mathcal{H}_{5}^{e1})\oplus \mathcal{R}_6. \\
		\end{split}  
\end{equation}
Next, we employ the dense connection to input features $\mathcal{H}_{i}^{sm}$, $\mathcal{H}_{i+1}^{sm}$, $\mathcal{H}_{i+2}^{sm}$, $\mathcal{H}_{i+3}^{sm}$ and $\mathcal{F}_{i}^{h}$ and obtain the predicted saliency map and the edge map with different levels through the use of the DJO decoder. This process can be expressed as follows:
\begin{equation}
		\begin{split}         
        &\mathcal{H}_{i}^{sm},\mathcal{H}_{i}^{em}  = \mathbb{DJO}(\mathcal{F}_{i}^{h},\mathcal{H}_{i+1}^{sm},\mathcal{H}_{i+2}^{sm},\mathcal{H}_{i+3}^{sm}), i=2,3,4 \\
		\end{split}  
\end{equation}
where $\mathbb{DJO(\cdot)}$ denotes the dual-branch joint optimization (DJO) decoder, and the index $i+1$, $i+2$, and $i+3$ are less than or equal to 5. The purpose of the DJO decoder is to seek and utilize the potential significant information of features with different layers, which can effectively distinguish differences between remote sensing objects and backgrounds.
\begin{table*}[t]
\renewcommand{\arraystretch}{1}
	\setlength{\tabcolsep}{0.5pt}
	\centering
	\caption{Attribute-based performance achieved on the challenging ORSI-4199 \cite{ORSI-4199}. ``$\uparrow$/$\downarrow$'' present that a higher/lower value is better. And top three scores in each like are marked in {\color{red} red}, {\color{green} green}, and {\color{blue} blue}.}
	\resizebox*{0.80\textwidth}{133mm}{
\begin{tabular}{cc|c|c|c|c|c|c|c|c|c|c|c|c|c|c|c|c|c}
\hline\hline
\multicolumn{2}{c|}{\textbf{Attribute}}                      & \textbf{CSNet$_{20}$}  & \textbf{VST$_{21}$}    & \textbf{DPORT$_{22}$} & \textbf{ICON$_{23}$} & \textbf{EMFI$_{22}$} & \textbf{CORR$_{22}$} & \textbf{ACCO$_{22}$} & \textbf{HFANet$_{22}$} & \textbf{ERPNet$_{22}$} & \textbf{MJRBM$_{22}$}  & \textbf{MCCNet$_{22}$} & \textbf{SeaNet$_{23}$} & \textbf{SRAL$_{23}$}   & \textbf{SASOD$_{23}$}  & \textbf{UG2L$_{23}$}   & \textbf{Ours-R} & \textbf{Ours-P} \\ \hline
\multicolumn{1}{c|}{\multirow{6}{*}{\textbf{SSO}}} & $\mathcal{M}$$\downarrow$ & 0.0205 & 0.0116 & 0.0147   & 0.0097   & 0.0117  & 0.0097  & 0.0125  & 0.0104 & 0.0117 & 0.0109 & 0.0103 & 0.0097 & \color{blue}0.0084 & 0.0105 & 0.0106 & \color{red}0.0078 & \color{green}0.0080 \\
\multicolumn{1}{c|}{}                     & $F_{\varphi}^{m}$$\uparrow$ & 0.7693 & 0.8080 & 0.7552   & 0.8218   & 0.8010  & 0.8190  & 0.8186  & 0.8237 & 0.8043 & 0.8100 & \color{blue}0.8286 & 0.8150 & 0.8186 & 0.8216 & 0.8195 & \color{green}0.8319 & \color{red}0.8333 \\
\multicolumn{1}{c|}{}                     & $F_{\varphi}^{m}$$\uparrow$ & 0.5237 & 0.6784 & 0.6853   & 0.7630   & 0.7549  & 0.7943  & 0.7523  & 0.7970 & 0.7123 & 0.6724 & \color{green} 0.7930 & 0.7791 & 0.7808 & 0.7784 & 0.7864 & \color{blue}0.7865 & \color{red}0.7931 \\
\multicolumn{1}{c|}{}                     & $F_{\varphi}^{w}$$\uparrow$ & 0.5297 & 0.7446 & 0.6987   & 0.7868   & 0.7572  & 0.7847  & 0.7798  & 0.7909 & 0.7406 & 0.7393 & \color{blue}0.7974 & 0.7821 & 0.7830 & 0.7813 & 0.7853 & \color{green}0.8019 & \color{red}0.8080 \\
\multicolumn{1}{c|}{}                     & $S_m$$\uparrow$  & 0.7727 & 0.8292 & 0.8017   & 0.8254   & 0.8261  & 0.8369  & \color{red}0.8461  & \color{green}0.8424 & 0.8305 & 0.8224 & 0.8392 & 0.8352 & 0.8372 & 0.8367 & 0.8414 & 0.8391 & \color{blue}0.8420 \\
\multicolumn{1}{c|}{}                     & $E_m$$\uparrow$  & 0.7724 & 0.8877 & 0.8888   & \color{blue}0.9501   & 0.9285  & 0.9486  & 0.9333  & 0.9473 & 0.9108 & 0.8868 & 0.9493 & 0.9438 & \color{green}0.9502 & 0.9377 & 0.9397 & \color{blue}0.9501 & \color{red}0.9552 \\ \hline
\multicolumn{1}{c|}{\multirow{6}{*}{\textbf{OC}}}  & $\mathcal{M}$$\downarrow$ & 0.0249 & 0.0133 & 0.0186   & 0.0153   & 0.0170  & 0.0155  & 0.0164  & \color{green}0.0125 & 0.0212 & 0.0145 & \color{red}0.0111 & 0.0135 & \color{blue}0.0123 & 0.0167 & 0.0154 & 0.0127 & 0.0130 \\
\multicolumn{1}{c|}{}                     & $F_{\varphi}^{m}$$\uparrow$ & 0.8118 & 0.8474 & 0.7800   & 0.8554   & 0.8223  & 0.8556  & 0.8478  & \color{blue}0.8617 & 0.8293 & 0.8465 & \color{red}0.8708 & 0.8454 & 0.8466 & 0.8473 & 0.8538 & 0.8603 & \color{green}0.8638 \\
\multicolumn{1}{c|}{}                     & $F_{\varphi}^{m}$$\uparrow$ & 0.6149 & 0.7401 & 0.7188   & 0.8051   & 0.7767  & 0.8356  & 0.7906  & \color{red}0.8409 & 0.7468 & 0.7250 & \color{green}0.8407 & 0.8105 & 0.8137 & 0.8030 & 0.8228 & 0.8201 & \color{blue}0.8285 \\
\multicolumn{1}{c|}{}                     & $F_{\varphi}^{w}$$\uparrow$ & 0.5954 & 0.7906 & 0.7253   & 0.8226   & 0.7740  & 0.8194  & 0.8074  & 0.8297 & 0.7644 & 0.7786 & \color{red}0.8407 & 0.8090 & 0.8119 & 0.8059 & 0.8155 & \color{blue}0.8298 & \color{green}0.8382 \\
\multicolumn{1}{c|}{}                     & $S_m$$\uparrow$  & 0.7953 & 0.8521 & 0.8115   & 0.8426   & 0.8305  & \color{blue}0.8530  & 0.8570  & \color{green}0.8602 & 0.8353 & 0.8407 & \color{red}0.8636 & 0.8479 & 0.8493 & 0.8475 & 0.8525 & 0.8469 & 0.8515 \\
\multicolumn{1}{c|}{}                     & $E_m$$\uparrow$  & 0.8263 & 0.9102 & 0.8935   & \color{blue}0.9552   & 0.9212  & 0.9539  & 0.9400  & 0.9555 & 0.9080 & 0.8990 & \color{red}0.9598 & 0.9459 & 0.9527 & 0.9373 & 0.9419 & 0.9534 & \color{green}0.9583 \\ \hline
\multicolumn{1}{c|}{\multirow{6}{*}{\textbf{NSO}}} & $\mathcal{M}$$\downarrow$ & 0.0429 & \color{blue}0.0182 & 0.0309   & 0.0225   & 0.0243  & 0.0248  & 0.0247  & 0.0206 & 0.0227 & 0.0326 & 0.0213 & 0.0196 & 0.0215 & 0.0209 & 0.0255 & \color{red}0.0164 & \color{green}0.0166 \\
\multicolumn{1}{c|}{}                     & $F_{\varphi}^{m}$$\uparrow$ & 0.8566 & \color{green}0.9109 & 0.8494   & 0.9025   & 0.8843  & 0.8932  & 0.8917  & 0.8980 & 0.8960 & 0.8742 & 0.8965 & 0.8988 & 0.8980 & 0.9038 & 0.8739 & \color{blue}0.9089 & \color{red}0.9162 \\
\multicolumn{1}{c|}{}                     & $F_{\varphi}^{m}$$\uparrow$ & 0.7955 & 0.8601 & 0.8130   & 0.8697   & 0.8534  & 0.8730  & 0.8508  & \color{blue}0.8804 & 0.8539 & 0.8169 & 0.8770 & 0.8784 & 0.8702 & 0.8789 & 0.8510 & \color{green}0.8832 & \color{red}0.8919 \\
\multicolumn{1}{c|}{}                     & $F_{\varphi}^{w}$$\uparrow$ & 0.7063 & 0.8685 & 0.7846   & \color{blue}0.8705   & 0.8430  & 0.8451  & 0.8527  & 0.8660 & 0.8404 & 0.8035 & 0.8641 & 0.8640 & 0.8605 & 0.8691 & 0.8299 & \color{green}0.8825 & \color{red}0.8929 \\
\multicolumn{1}{c|}{}                     & $S_m$$\uparrow$  & 0.8252 & \color{red}0.8878 & 0.8355   & 0.8731   & 0.8650  & 0.8660  & 0.8792  & 0.8805 & 0.8725 & 0.8511 & 0.8799 & 0.8745 & 0.8775 & \color{blue}0.8813 & 0.8608 & 0.8801 & \color{green}0.8841 \\
\multicolumn{1}{c|}{}                     & $E_m$$\uparrow$  & 0.9187 & 0.9577 & 0.9102   & 0.9582   & 0.9484  & 0.9491  & 0.9516  & 0.9615 & 0.9496 & 0.9347 & 0.9598 & \color{blue}0.9623 & 0.9577 & 0.9622 & 0.9410 & \color{green}0.9630 & \color{red}0.9660 \\ \hline
\multicolumn{1}{c|}{\multirow{6}{*}{\textbf{MSO}}} & $\mathcal{M}$$\downarrow$ & 0.0360 & \color{blue}0.0180 & 0.0282   & 0.0196   & 0.0223  & 0.0219  & 0.0191  & 0.0209 & 0.0234 & 0.0222 & 0.0191 & 0.0205 & 0.0192 & 0.0182 & 0.0199 & \color{green}0.0166 & \color{red}0.0153 \\
\multicolumn{1}{c|}{}                     & $F_{\varphi}^{m}$$\uparrow$ & 0.8216 & 0.8649 & 0.8260   & 0.8637   & 0.8589  & 0.8671  & 0.8706  & 0.8676 & 0.8568 & 0.8591 & 0.8718 & 0.8616 & 0.8657 & \color{blue}0.8725 & 0.8700 & \color{green}0.8741 & \color{red}0.8794 \\
\multicolumn{1}{c|}{}                     & $F_{\varphi}^{m}$$\uparrow$ & 0.6710 & 0.7711 & 0.7729   & 0.8185   & 0.8196  & \color{red}0.8466  & 0.8198  & 0.8385 & 0.7860 & 0.7709 & 0.8437 & 0.8346 & 0.8325 & 0.8373 & \color{blue}0.8440 & 0.8418 & \color{green}0.8452 \\
\multicolumn{1}{c|}{}                     & $F_{\varphi}^{w}$$\uparrow$ & 0.6501 & 0.8112 & 0.7727   & 0.8272   & 0.8190  & 0.8257  & 0.8328  & 0.8312 & 0.7960 & 0.7929 & \color{blue}0.8388 & 0.8278 & 0.8305 & 0.8359 & 0.8387 & \color{green}0.8484 & \color{red}0.8554 \\
\multicolumn{1}{c|}{}                     & $S_m$$\uparrow$  & 0.8082 & 0.8636 & 0.8340   & 0.8565   & 0.8651  & 0.8633  &  \color{green} 0.8789  & 0.8702 & 0.8614 & 0.8511 & 0.8675 & 0.8638 & 0.8689 & 0.8706 & \color{blue}0.8751 & 0.8737 & \color{red}0.8795 \\
\multicolumn{1}{c|}{}                     & $E_m$$\uparrow$  & 0.8555 & 0.9207 & 0.9189   & \color{blue}0.9560   & 0.9419  & 0.9491  & 0.9493  & 0.9467 & 0.9280 & 0.9219 & 0.9546 & 0.9513 & 0.9525 & 0.9516 & 0.9507 & \color{green}0.9597 & \color{red}0.9634 \\ \hline
\multicolumn{1}{c|}{\multirow{6}{*}{\textbf{LSO}}} & $\mathcal{M}$$\downarrow$ & 0.0354 & 0.0182 & 0.0440   & \color{blue}0.0177   & 0.0207  & 0.0236  & 0.0239  & 0.0210 & 0.0242 & 0.0258 & 0.0242 & 0.0186 & 0.0201 & 0.0200 & 0.0210 & \color{green}0.0158 & \color{red}0.0136 \\
\multicolumn{1}{c|}{}                     & $F_{\varphi}^{m}$$\uparrow$ & 0.7782 & 0.8183 & 0.7304   & \color{blue}0.8251   & 0.8104  & 0.7953  & 0.8117  & 0.8109 & 0.8004 & 0.7992 & 0.8056 & 0.8138 & 0.8104 & 0.8151 & 0.8090 & \color{green}0.8273 & \color{red}0.8345 \\
\multicolumn{1}{c|}{}                     & $F_{\varphi}^{m}$$\uparrow$ & 0.6542 & 0.7452 & 0.6807   & \color{blue}0.7940   & 0.7719  & 0.7802  & 0.7710  & 0.7811 & 0.7500 & 0.7323 & 0.7857 & 0.7922 & 0.7696 & 0.7769 & 0.7809 & \color{green}0.8014 & \color{red}0.8114 \\
\multicolumn{1}{c|}{}                     & $F_{\varphi}^{w}$$\uparrow$ & 0.6156 & 0.7748 & 0.6572   & \color{blue}0.7934   & 0.7746  & 0.7597  & 0.7764  & 0.7790 & 0.7517 & 0.7396 & 0.7756 & 0.7815 & 0.7745 & 0.7768 & 0.7703 & \color{green}0.8019 & \color{red}0.8121 \\
\multicolumn{1}{c|}{}                     & $S_m$$\uparrow$  & 0.8009 & \color{blue}0.8504 & 0.7680   & 0.8503   & 0.8401  & 0.8290  & 0.8450  & 0.8425 & 0.8342 & 0.8283 & 0.8375 & 0.8420 & 0.8404 & 0.8425 & 0.8371 & \color{green}0.8549 & \color{red}0.8614 \\
\multicolumn{1}{c|}{}                     & $E_m$$\uparrow$  & 0.8351 & 0.9166 & 0.8654   & \color{blue}0.9544   & 0.9342  & 0.9406  & 0.9343  & 0.9379 & 0.9167 & 0.9095 & 0.9419 & 0.9509 & 0.9360 & 0.9372 & 0.9332 & \color{green}0.9555 & \color{red}0.9614 \\ \hline
\multicolumn{1}{c|}{\multirow{6}{*}{\textbf{ISO}}} & $\mathcal{M}$$\downarrow$ & 0.1281 & \color{green}0.0588 & 0.1555   & \color{blue} 0.0604   & 0.0822  & 0.1014  & 0.0739  & 0.0781 & 0.0955 & 0.0865 & 0.0766 & 0.0793 & 0.0770 & 0.0718 & 0.0782 & 0.0656 & \color{red}0.0538 \\
\multicolumn{1}{c|}{}                     & $F_{\varphi}^{m}$$\uparrow$ & 0.8731 & \color{blue}0.9346 & 0.8191   & \color{green}0.9355   & 0.9149  & 0.8979  & 0.9159  & 0.9183 & 0.8990 & 0.9024 & 0.9114 & 0.9163 & 0.9141 & 0.9219 & 0.9170 & 0.9217 & \color{red}0.9369 \\
\multicolumn{1}{c|}{}                     & $F_{\varphi}^{m}$$\uparrow$ & 0.7669 & 0.7890 & 0.7613   & \color{green}0.9090   & 0.7824  & 0.8702  & 0.8755  & 0.7707 & 0.7916 & 0.8733 & 0.8923 & 0.8963 & 0.7550 & 0.8183 & 0.8618 & \color{blue}0.9085 & \color{red}0.9256 \\
\multicolumn{1}{c|}{}                     & $F_{\varphi}^{w}$$\uparrow$ & 0.7482 & \color{green}0.8939 & 0.7099   & 0.8954   & 0.8600  & 0.8213  & 0.8634  & 0.8650 & 0.8250 & 0.8417 & 0.8616 & 0.8625 & 0.8611 & 0.8728 & 0.8614 & \color{blue}0.8849 & \color{red}0.9062 \\
\multicolumn{1}{c|}{}                     & $S_m$$\uparrow$  & 0.7993 & \color{green}0.8819 & 0.7306   & 0.8792   & 0.8486  & 0.8191  & 0.8587  & 0.8530 & 0.8274 & 0.8425 & 0.8520 & 0.8502 & 0.8509 & 0.8630 & 0.8516 & \color{blue}0.8674 & \color{red}0.8860 \\
\multicolumn{1}{c|}{}                     & $E_m$$\uparrow$  & 0.7728 & 0.8290 & 0.7493   & \color{blue}0.8944   & 0.8005  & 0.8500  & 0.8538  & 0.7993 & 0.7948 & 0.8658 & 0.8854 & 0.8798 & 0.7917 & 0.8113 & 0.8372 & \color{green}0.9050 & \color{red}0.9266 \\ \hline
\multicolumn{1}{c|}{\multirow{6}{*}{\textbf{CSO}}} & $\mathcal{M}$$\downarrow$ & 0.0701 & 0.0390 & 0.0776   & \color{blue}0.0384   & 0.0458  & 0.0514  & 0.0432  & 0.0444 & 0.0484 & 0.0513 & 0.0435 & 0.0434 & 0.0424 & 0.0408 & 0.0449 & \color{green}0.0378 & \color{red}0.0325 \\
\multicolumn{1}{c|}{}                     & $F_{\varphi}^{m}$$\uparrow$ & 0.8674 & 0.9104 & 0.8351   & \color{green}0.9165   & 0.8991  & 0.8957  & 0.9052  & 0.9040 & 0.8961 & 0.8921 & 0.9029 & 0.9047 & 0.9072 & 0.9114 & 0.9010 & \color{blue}0.9159 & \color{red}0.9243 \\
\multicolumn{1}{c|}{}                     & $F_{\varphi}^{m}$$\uparrow$ & 0.7725 & 0.8304 & 0.7828   & \color{blue}0.8916   & 0.8392  & 0.8761  & 0.8685  & 0.8373 & 0.8302 & 0.8496 & 0.8864 & 0.8853 & 0.8359 & 0.8598 & 0.8689 & \color{green}0.8988 & \color{red}0.9085 \\
\multicolumn{1}{c|}{}                     & $F_{\varphi}^{w}$$\uparrow$ & 0.7398 & 0.8667 & 0.7614   & \color{blue}0.8802   & 0.8569  & 0.8445  & 0.8621  & 0.8625 & 0.8421 & 0.8350 & 0.8653 & 0.8649 & 0.8673 & 0.8715 & 0.8587 & \color{green}0.8851 & \color{red}0.8981 \\
\multicolumn{1}{c|}{}                     & $S_m$$\uparrow$  & 0.8430 & 0.8903 & 0.8090   & \color{blue}0.8903   & 0.8760  & 0.8645  & 0.8851  & 0.8803 & 0.8726 & 0.8690 & 0.8804 & 0.8799 & 0.8825 & 0.8882 & 0.8791 & \color{green}0.8917 & \color{red}0.9014 \\
\multicolumn{1}{c|}{}                     & $E_m$$\uparrow$  & 0.8489 & 0.8992 & 0.8485   & \color{blue}0.9339   & 0.8932  & 0.9115  & 0.9121  & 0.8921 & 0.8868 & 0.9077 & 0.9272 & 0.9273 & 0.8963 & 0.9014 & 0.9076 & \color{green}0.9384 & \color{red}0.9488 \\ \hline
\multicolumn{1}{c|}{\multirow{6}{*}{\textbf{CS}}}  & $\mathcal{M}$$\downarrow$ & 0.0682 & 0.0383 & 0.0752   & \color{blue}0.0374   & 0.0438  & 0.0485  & 0.0417  & 0.0425 & 0.0459 & 0.0488 & 0.0410 & 0.0421 & 0.0404 & 0.0393 & 0.0433 & \color{green}0.0367 & \color{red}0.0316 \\
\multicolumn{1}{c|}{}                     & $F_{\varphi}^{m}$$\uparrow$ & 0.8725 & 0.9145 & 0.8399   & \color{green}0.9211   & 0.9038  & 0.9019  & 0.9103  & 0.9090 & 0.9015 & 0.8976 & 0.9092 & 0.9099 & 0.9129 & 0.9166 & 0.9058 & \color{blue}0.9201 & \color{red}0.9293 \\
\multicolumn{1}{c|}{}                     & $F_{\varphi}^{m}$$\uparrow$ & 0.7762 & 0.8383 & 0.7888   & \color{blue}0.8959   & 0.8462  & 0.8825  & 0.8735  & 0.8442 & 0.8357 & 0.8544 & 0.8925 & 0.8899 & 0.8445 & 0.8653 & 0.8743 & \color{green}0.9024 & \color{red}0.9131 \\
\multicolumn{1}{c|}{}                     & $F_{\varphi}^{w}$$\uparrow$ & 0.7423 & 0.8699 & 0.7673   & \color{blue}0.8842   & 0.8621  & 0.8514  & 0.8668  & 0.8676 & 0.8476 & 0.8407 & 0.8717 & 0.8695 & 0.8725 & 0.8765 & 0.8634 & \color{green}0.8885 & \color{red}0.9028 \\
\multicolumn{1}{c|}{}                     & $S_m$$\uparrow$  & 0.8459 & 0.8926 & 0.8134   & \color{blue}0.8932   & 0.8805  & 0.8698  & 0.8886  & 0.8844 & 0.8772 & 0.8738 & 0.8855 & 0.8834 & 0.8863 & 0.8916 & 0.8830 & \color{green}0.8941 & \color{red}0.9046 \\
\multicolumn{1}{c|}{}                     & $E_m$$\uparrow$  & 0.8517 & 0.9041 & 0.8535   & \color{blue}0.9371   & 0.8986  & 0.9166  & 0.9165  & 0.8972 & 0.8918 & 0.9115 & 0.9319 & 0.9310 & 0.9020 & 0.9065 & 0.9125 & \color{green}0.9407 & \color{red}0.9514 \\ \hline
\multicolumn{1}{c|}{\multirow{6}{*}{\textbf{BSO}}} & $\mathcal{M}$$\downarrow$ & 0.1043 & \color{blue}0.0568 & 0.1303   & \color{blue}0.0568   & 0.0673  & 0.0802  & 0.0621  & 0.0655 & 0.0729 & 0.0757 & 0.0638 & 0.0657 & 0.0629 & 0.0610 & 0.0647 & \color{green}0.0566 & \color{red}0.0473 \\
\multicolumn{1}{c|}{}                     & $F_{\varphi}^{m}$$\uparrow$ & 0.8991 & 0.9371 & 0.8438   & \color{green}0.9405   & 0.9321  & 0.9200  & 0.9337  & 0.9331 & 0.9259 & 0.9143 & 0.9301 & 0.9314 & 0.9334 & 0.9341 & 0.9325 & \color{blue}0.9392 & \color{red}0.9489 \\
\multicolumn{1}{c|}{}                     & $F_{\varphi}^{m}$$\uparrow$ & 0.8162 & 0.8272 & 0.7804   & \color{blue}0.9219   & 0.8247  & 0.8978  & 0.9028  & 0.8056 & 0.8318 & 0.8957 & 0.9151 & 0.9149 & 0.8031 & 0.8587 & 0.8934 & \color{green}0.9277 & \color{red}0.9397 \\
\multicolumn{1}{c|}{}                     & $F_{\varphi}^{w}$$\uparrow$ & 0.8024 & 0.8962 & 0.7512   & \color{blue}0.8988   & 0.8839  & 0.8566  & 0.8860  & 0.8852 & 0.8687 & 0.8569 & 0.8853 & 0.8838 & 0.8882 & 0.8915 & 0.8849 & \color{green}0.9024 & \color{red}0.9194 \\
\multicolumn{1}{c|}{}                     & $S_m$$\uparrow$  & 0.8423 & \color{green}0.8937 & 0.7714   & \color{blue}0.8927   & 0.8754  & 0.8546  & 0.8849  & 0.8796 & 0.8694 & 0.8636 & 0.8794 & 0.8771 & 0.8813 & 0.8873 & 0.8789 & 0.8904 & \color{red}0.9054 \\
\multicolumn{1}{c|}{}                     & $E_m$$\uparrow$  & 0.8186 & 0.8532 & 0.7830   & \color{blue}0.9094   & 0.8373  & 0.8826  & 0.8826  & 0.8287 & 0.8334 & 0.8881 & 0.9078 & 0.9045 & 0.8305 & 0.8475 & 0.8724 & \color{green}0.9205 & \color{red}0.9364 \\ \hline\hline
\end{tabular}}

\label{table2}
\end{table*}
\subsection{Loss Functions}
    In the proposed UDCNet model, we supervise multi-level salient features $\{\mathcal{H}_{i}^{sm}\}_{i=2}^{5}$, $\mathcal{R}_{6}$, and edge features $\{\mathcal{H}_{i}^{em}\}_{i=2}^{5}$ to optimize model parameters for generating high-quality predicted map. Specifically, we employ the weighted binary cross-entropy (BCE) function and the weighted intersection over union (IoU) function for the salient segmentation supervision. We utilize the dice loss function to help learn complete boundaries and improve the accuracy of segmentation in the edge supervision. It is worth noting that all predicted maps will be uniformly resized to the same size as the ground truths using bilinear interpolation during the supervision process. The total loss function is formulated as follows:
\begin{equation}
		\begin{split}         
        &\mathcal{L}_{total}=\frac{1}{2^{4}}(\mathcal{L}_{bce}^{w}(\mathcal{R}_6,G_s)\oplus \mathcal{L}_{iou}^{w}(\mathcal{R}_6,G_s)) \\ 
       &\quad\quad\quad \oplus \sum_{i=2}^{5}\frac{1}{2^{i-2}}(\mathcal{L}_{bce}^{w}(\mathcal{H}_i^{sm},G_s)\oplus \mathcal{L}_{iou}^{w}(\mathcal{H}_i^{sm},G_s)) \\
        &\quad\quad\quad \oplus \sum_{i=2}^{5}\frac{1}{2^{i-1}}(\mathcal{L}_{dice}(\mathcal{H}_i^{em},G_e)), i = 2,3,4,5 \\
		\end{split}  
\end{equation}
where $\mathcal{L}_{bce}^{w}(\cdot,\cdot)$ and $\mathcal{L}_{iou}^{w}(\cdot,\cdot)$ present the weighted binary cross-entropy (BCE) and weighted intersection over union (IoU) functions, $\mathcal{L}_{dice}(\cdot,\cdot)$ denotes the dice loss function. $G_s$ and $G_e$ are the saliency ground truth and the edge ground truth, respectively. 

\section{Experiments}
\subsection{Experimental Settings}
\subsubsection{Datasets}
We evaluate the proposed UDCNet model on three widely-used ORSIs-SOD datasets, including ORSSD \cite{ORSSD}, EORSSD \cite{EORSSD}, and ORSI-4199 \cite{ORSI-4199}. ORSSD \cite{ORSSD} contains 800 images with remote sensing objects, comprising 600 training images and 200 testing images. EORSSD \cite{EORSSD} is an extended version of ORSSD, including 1400 training images and 600 testing images. ORSI-4199 \cite{ORSI-4199} is a large-scale dataset with abundant classes of remote sensing objects, which can be divided into a training set of 2000 samples and a testing set of 2199 samples. In a similar manner to \cite{MCCNet,AESINet}, we train the proposed UDCNet method using the training set of each dataset and evaluate its performance on the testset.

\subsubsection{Implementation details}
We implement the proposed UDCNet model using the PyTorch framework and train it on four NVIDIA GTX 4090 GPUs with 24 GB memory. For the backbone, we adopt either pre-trained ResNet50 \cite{ResNet} or PVTv2 \cite{Pvt2}. The optimizer is Adam, with an initial learning rate of 1e-4 and a decay rate of 0.1 every 60 epochs. During the training process, we set the input image size to 352$\times$352 and the batch size to 40. The entire training process is iterated for 180 epochs. Following \cite{MCCNet,ACCorNet,AESINet}, we use data augmentation techniques such as horizontal flipping, rotation, and random cropping to enhance the robustness of the model.

\subsubsection{Evaluation Metrics}
We employ some widely-adopted evaluation metrics to evaluate the effectiveness of our UDCNet method, including the mean absolute error ($\mathcal{M}$), maximum F-measure ($F_{\varphi}^{m}$), average F-measure ($F_{\varphi}^{a}$), weight F-measure ($F_{\varphi}^{w}$), S-measure ($S_m$)\nocite{Sm}, and E-measure ($E_m$). Among the metrics, a smaller value for $\mathcal{M}$ indicates better performance, while larger values for $F_{\varphi}^{m}$, $F_{\varphi}^{a}$, $F_{\varphi}^{w}$, $S_m$, and $E_m$ indicate better performance. In addition, we provide the precision-recall ($PR$) and F-measure ($F_m$) curves to further validate the performance of the proposed UDCNet method.

\subsection{Comparison with the State-of-the-Art Methods}
We compare our UDCNet model with 24 state-of-the-art NSIs-SOD methods and ORSIs-SOD methods on three public datasets, including ten NSIs-SOD methods ($i.e.$, MINet \cite{MINet}, ITSD \cite{ITSD}, CSNet \cite{CSF}, GateNet \cite{GateNet}, PA-KRN \cite{PAKRN}, SAMNet \cite{SAMNet}, VST \cite{VST}, DRORTNet \cite{DPORTNet}, DNTD \cite{DNRF}, and ICON-P \cite{ICON}) and fourteen ORSIs-SOD methods ($i.e.,$, LVNet \cite{ORSSD}, DAFNet \cite{EORSSD}, EMFNet \cite{EMFINet}, EPRNet \cite{ERPNet}, HFANet \cite{HFANet}, MJRBM \cite{ORSI-4199}, ACCoNet \cite{ACCorNet}, CorrNet \cite{CoorNet}, MCCNet \cite{MCCNet}, SeaNet \cite{SeaNet}, AESINet \cite{AESINet}, SRAL \cite{SRAL}, SASOD \cite{SASOD}, and UG2L \cite{UG2L}). Note that all the prediction results are obtained through open-source codes or directly provided by the authors.

\begin{table*}[t]
\renewcommand{\arraystretch}{1}
	\setlength{\tabcolsep}{2.5pt}
	\centering
	\caption{Ablation analysis of our UDCNet method structure.}
	\resizebox{0.88\textwidth}{20mm}{
\begin{tabular}{ccccc|cccccc|cccccc}
\hline\hline
\multicolumn{5}{c|}{Structure Settings}                                                                                           & \multicolumn{6}{c|}{ORSSD (200 images)}              & \multicolumn{6}{c}{EORSSD (600 images)}              \\ \cline{1-5}
\rowcolor{red!15}\multicolumn{1}{c|}{Num.} & \multicolumn{1}{c|}{Baseline} & \multicolumn{1}{c|}{FSDT} & \multicolumn{1}{c|}{DJO} & DSE  &\cellcolor{blue!15}$\mathcal{M}$$\downarrow$   & \cellcolor{blue!15}$F_{\varphi}^{m}$$\uparrow$   & \cellcolor{blue!15}$F_{\varphi}^{a}$$\uparrow$   & \cellcolor{blue!15}$F_{\varphi}^{w}$$\uparrow$   & \cellcolor{blue!15}$S_m$$\uparrow$     & \cellcolor{blue!15}$E_m$$\uparrow$     & \cellcolor{blue!15}$\mathcal{M}$$\downarrow$   & \cellcolor{blue!15}$F_{\varphi}^{m}$$\uparrow$   & \cellcolor{blue!15}$F_{\varphi}^{a}$$\uparrow$   & \cellcolor{blue!15}$F_{\varphi}^{w}$$\uparrow$   & \cellcolor{blue!15}$S_m$$\uparrow$     & \cellcolor{blue!15}$E_m$$\uparrow$     \\ \hline
\multicolumn{1}{c|}{(a)}    & \multicolumn{1}{c|}{$\checkmark$}        & \multicolumn{1}{c|}{}        & \multicolumn{1}{c|}{}        &         & 0.0143 & 0.8641 & 0.7780 & 0.8144 & 0.8751 & 0.9218 & 0.0124 & 0.7923 & 0.6470 & 0.7182 & 0.8128 & 0.8492 \\ 
\multicolumn{1}{c|}{(b)}    & \multicolumn{1}{c|}{$\checkmark$}        & \multicolumn{1}{c|}{$\checkmark$}       & \multicolumn{1}{c|}{}        &         & 0.0078 & 0.9148 & 0.8605 & 0.8882 & 0.9258 & 0.9710 & 0.0061 & 0.8844 & 0.7937 & 0.8479 & 0.8814 & 0.9460 \\ 
\multicolumn{1}{c|}{(c)}    & \multicolumn{1}{c|}{$\checkmark$}        & \multicolumn{1}{c|}{}        & \multicolumn{1}{c|}{$\checkmark$}       &         & 0.0070 & 0.9212 & 0.8827 & 0.9013 & 0.9297 & 0.9760 & 0.0066 & 0.8697 & 0.7769 & 0.8306 & 0.8830 & 0.9326 \\ 
\multicolumn{1}{c|}{(d)}    & \multicolumn{1}{c|}{$\checkmark$}        & \multicolumn{1}{c|}{}        & \multicolumn{1}{c|}{}        & $\checkmark$       & 0.0127 & 0.8907 & 0.8081 & 0.8405 & 0.8953 & 0.9394 & 0.0080 & 0.8767 & 0.7926 & 0.8390 & 0.8803 & 0.9407 \\ 
\multicolumn{1}{c|}{(e)}    & \multicolumn{1}{c|}{$\checkmark$}        & \multicolumn{1}{c|}{$\checkmark$}       & \multicolumn{1}{c|}{$\checkmark$}       &         & \textbf{\color{red}0.0066} & 0.9258 & 0.8896 & 0.9077 & 0.9362 & \textbf{\color{red}0.9771} & 0.0060 & 0.8871 & 0.8167 & 0.8598 & 0.8950 & 0.9539 \\ 
\multicolumn{1}{c|}{(f)}    & \multicolumn{1}{c|}{$\checkmark$}        & \multicolumn{1}{c|}{$\checkmark$}       & \multicolumn{1}{c|}{}        & $\checkmark$       & 0.0070 & 0.9198 & 0.8771 & 0.8981 & 0.9315 & 0.9758 & 0.0062 & 0.8846 & 0.8121 & 0.8553 & 0.8885 & 0.9519 \\ 
\multicolumn{1}{c|}{(g)}    & \multicolumn{1}{c|}{$\checkmark$}        & \multicolumn{1}{c|}{}        & \multicolumn{1}{c|}{$\checkmark$}       & $\checkmark$       & 0.0069 & 0.9241 & 0.8930 & 0.9092 & 0.9370 & 0.9831 & 0.0061 & 0.8855 & \textbf{\color{red}0.8267} & 0.8604 & 0.8906 & 0.9543 \\ \hline
\multicolumn{1}{c|}{(h)}    & \multicolumn{1}{c|}{$\checkmark$}        & \multicolumn{1}{c|}{$\checkmark$}       & \multicolumn{1}{c|}{$\checkmark$}       & $\checkmark$       & 0.0068 & \textbf{\color{red}0.9267} & \textbf{\color{red}0.8932} & \textbf{\color{red}0.9093} & \textbf{\color{red}0.9389} & 0.9770 & \textbf{\color{red}0.0056} & \textbf{\color{red}0.8903} & 0.8211 & \textbf{\color{red}0.8615} & \textbf{\color{red}0.8957} & \textbf{\color{red}0.9547} \\ \hline\hline
\end{tabular}}
\label{table3}
\end{table*}
\begin{figure}[t]
    \centering\includegraphics[width=0.46\textwidth,height=3.5cm]{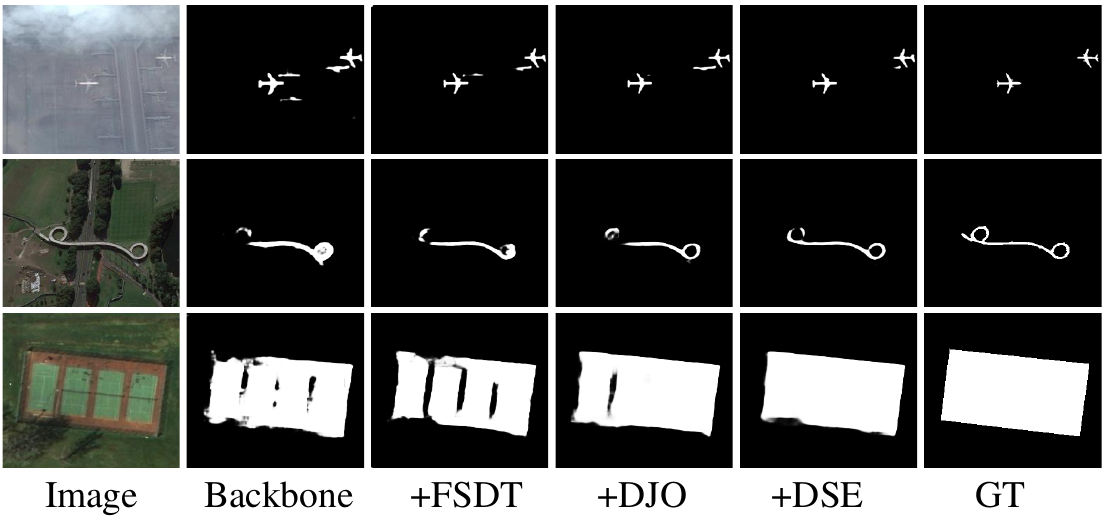}
    \captionsetup{font={small}, justification=raggedright}
    \caption{ Visual results of the effectiveness of our modules.}
    \label{Fig.8}
\end{figure}
\begin{table}[t]
\renewcommand{\arraystretch}{1}
	\setlength{\tabcolsep}{2pt}
	\centering
	\caption{Ablation analysis of our FSDT block structure.}
	\resizebox{0.48\textwidth}{9mm}{
\begin{tabular}{c|cccccc|cccccc}
 \hline \hline
\multirow{2}{*}{Method} & \multicolumn{6}{c|}{ORSSD (200 images)}              & \multicolumn{6}{c}{EORSSD (600 images)}              \\ 
                        &\cellcolor{blue!15}$\mathcal{M}$$\downarrow$   & \cellcolor{blue!15}$F_{\varphi}^{m}$$\uparrow$   & \cellcolor{blue!15}$F_{\varphi}^{a}$$\uparrow$   & \cellcolor{blue!15}$F_{\varphi}^{w}$$\uparrow$   & \cellcolor{blue!15}$S_m$$\uparrow$     & \cellcolor{blue!15}$E_m$$\uparrow$     & \cellcolor{blue!15}$\mathcal{M}$$\downarrow$   & \cellcolor{blue!15}$F_{\varphi}^{m}$$\uparrow$   & \cellcolor{blue!15}$F_{\varphi}^{a}$$\uparrow$   & \cellcolor{blue!15}$F_{\varphi}^{w}$$\uparrow$   & \cellcolor{blue!15}$S_m$$\uparrow$     & \cellcolor{blue!15}$E_m$$\uparrow$     \\ \hline
FSDT W/Fre             & 0.0093 & 0.8993 & 0.8443 & 0.8712 & 0.9033 & 0.9601 & 0.0069 & 0.8777 & 0.7703 & 0.8320 & 0.8824 & 0.9337 \\ 
FSDT W/Spa             & 0.0104 & 0.9066 & 0.8370 & 0.8673 & 0.9028 & 0.9570 & 0.0069 & 0.8778 & 0.7880 & 0.8408 & 0.8856 & 0.9424 \\ 
FSDT W/o AFS         & 0.0090 & 0.9137 & 0.8492 & \textbf{\color{red}0.8830} & 0.9202 & 0.9646 & 0.0064 & \textbf{\color{red}0.8886} & 0.7855 & 0.8448 & \textbf{\color{red}0.8819} & 0.9411 \\ \hline
FSDT block                 & \textbf{\color{red}0.0078} & \textbf{\color{red}0.9148} & \textbf{\color{red}0.8605} & 0.8882 & \textbf{\color{red}0.9258} & \textbf{\color{red}0.9710} & \textbf{\color{red}0.0061} & 0.8844 & \textbf{\color{red}0.7937} & \textbf{\color{red}0.8479} & 0.8814 & \textbf{\color{red}0.9460} \\ \hline\hline
\end{tabular}}
\label{table4}
\end{table}
\begin{figure}[t]
\centering\includegraphics[width=0.48\textwidth,height=3.8cm]{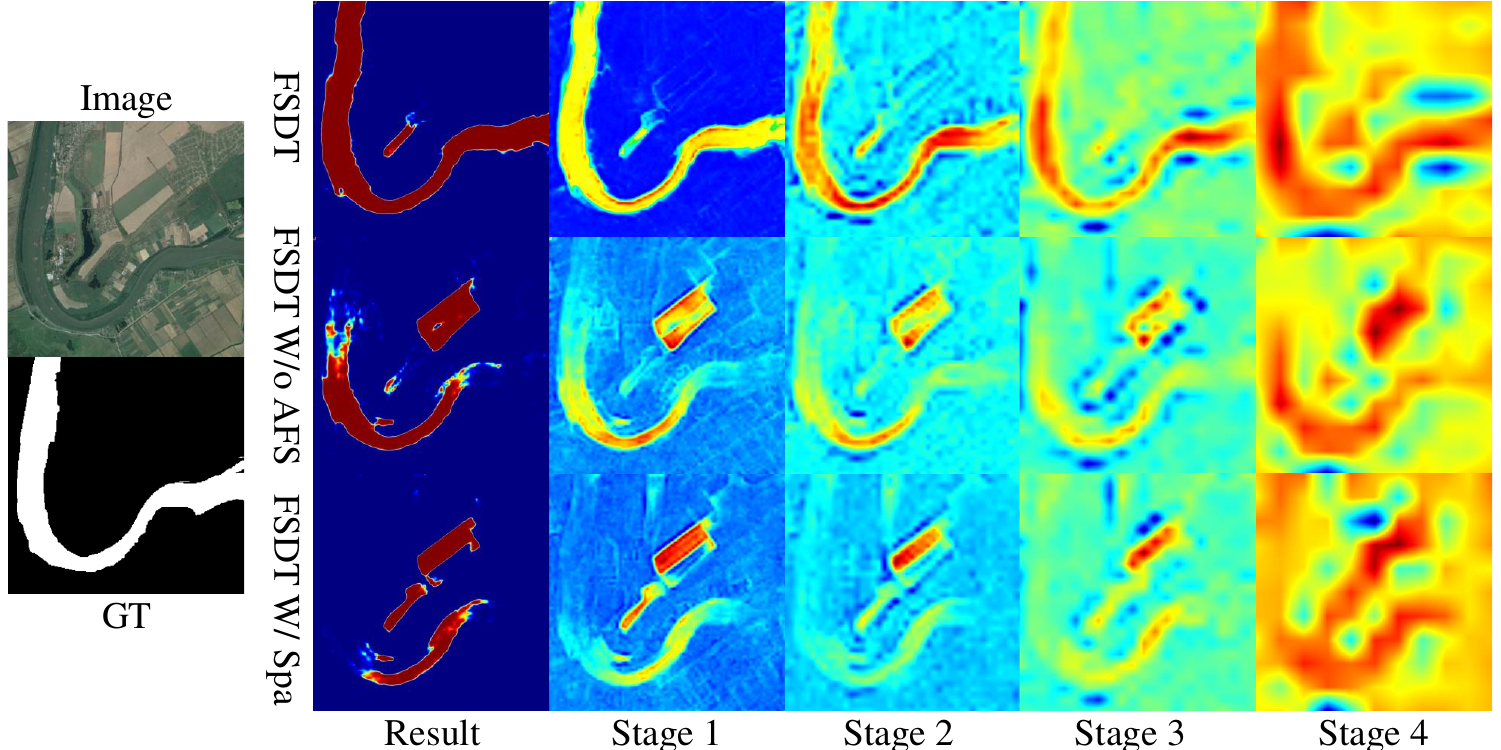}
	\captionsetup{font={small}, justification=raggedright}
	\caption{ Visual comparisons of the proposed FSDT block.}
	\label{Fig.9}

\end{figure}
\subsubsection{Quantitative comparisons}
Table \ref{table1} provides a comprehensive overview of the quantitative results of the proposed UDCNet method in comparison to 24 state-of-the-art SOD methods across three public remote sensing datasets. From Table \ref{table1}, it can be seen that our UDCNet method achieves outstanding performance, irrespective of the type of backbone network used. Specifically, our UDCNet method with ResNet50 \cite{ResNet} backbone overall surpasses the recently proposed UG2L \cite{UG2L} methods by 44.11\%, 23.21\%, and 19.54\% in terms of $\mathcal{M}$ on ORSSD \cite{ORSSD}, EORSSD \cite{EORSSD}, and ORSI-4199 \cite{ORSI-4199} datasets. Similarly, there is a significant improvement in performance across other metrics. In addition, compared to the Transformer backbone-based methods ($i.e.$, ICON-P \cite{ICON} and HFANet \cite{HFANet}), the proposed UDCNet method with PVTv2 \cite{Pvt2} backbone achieves remarkable performance boost of 18.99\%, 1.03\%, 2.32\%, 2.21\%, 1.37\%, and 1.10\% over ICON-P \cite{ICON} model and 32.49\%, 1.61\%, 4.88\%, 3.03\%, 1.25\% and 3.87\% over HFANet \cite{HFANet} model in terms of six evaluation metrics ($i.e.$, $\mathcal{M}$, $F_{\varphi}^{m}$, $F_{\varphi}^{a}$, $F_{\varphi}^{w}$, $S_m$, and $E_m$ ) on ORSI-4199 \cite{ORSI-4199} dataset. Likewise, substantial improvements in performance are observed on ORSSD \cite{ORSSD} and EORSSD \cite{EORSSD} datasets. Furthermore, in Fig. \ref{Fig.5}, we present the $PR$ and $F_m$ curves of our UDCNet and several SOTA models on three public datasets. Our UDCNet method is highlighted in {\color{red}red}. As can be seen from the curves, the proposed UDCNet method outperforms some competitive methods with a clear advantage. In conclusion, these quantitative results provide powerful evidence that the UDCNet method exhibits higher accuracy and effectiveness. This is thanks to the exploration of spectral features from frequency domain by our UDCNet.

\subsubsection{Qualitative comparisons}
Fig. \ref{Fig.6} gives the visual result of the proposed UDCNet method with 11 state-of-the-art ORSIs-SOD methods under some typical remote sensing scenarios, including ship ($1^{st}$ row), river ($2^{nd}$ row), road ($3^{rd}$ row), stadium ($4^{th}$ row), and some residential buildings ($5^{th}$-$7^{th}$ rows). By observing the predicted results in Fig. \ref{Fig.6}, it can be seen that our UDCNet method achieves precise segmentation for remote sensing objects in various scenes, whereas the recent spatial domain-based methods ($e.g.$, UG2L \cite{UG2L}, SASOD \cite{SASOD}, and SRAL \cite{SRAL}) fail to accomplish this level of accuracy. This is attributed to the fact that certain positive information in the frequency domain can enhance pixels from the spatial domain.
\subsubsection{Attribute comparisons}
To further analyze the performance of our model, we conduct an attribute analysis on the challenging ORSI-4199 \cite{ORSI-4199} dataset in Table \ref{table2}, which includes 9 attributes, that is, small salient object (SSO), off center (OC), narrow salient object (NSO), multiple salient objects (MSO), low contrast scene (LCS), incomplete salient object (ISO), complex salient object (CSO), complex scene (CS), and big salient objects (BSO). We compare the proposed UDCNet method with 15 state-of-the-art methods ($i.e.$, CSNet \cite{CSF}, VST \cite{VST}, DPORT \cite{DPORTNet}, ICON \cite{ICON}, EMFI \cite{EMFINet}, CORR \cite{CoorNet}, ACCO \cite{ACCorNet}, HFANet \cite{HFANet}, ERPNet \cite{ERPNet}, MJRBM \cite{ORSI-4199}, MCCNet \cite{MCCNet}, SeaNet \cite{SeaNet}, SRAL \cite{SRAL}, SASOD \cite{SASOD}, and UG2L \cite{UG2L}). From Table \ref{table2}, our UDCNet method achieves excellent results in most challenging attributes, whether based on the ResNet50 \cite{ResNet} or PVTv2 \cite{Pvt2} backbone, demonstrating the robustness and efficiency of the proposed UDCNet approach. The superiority in performance benefits from the joint optimization of the FSDT block, DSE module, and DJO decoder for initial input features from encoder.
\subsubsection{Efficiency analysis} 
Fig. \ref{Fig.7} shows the FLOPs and parameters comparison between our UDCNet and existing SOTA methods. To ensure a fair comparison, all methods are performed on a single 4090 GPU with 24 GB memory, and input image size set to 352 $\times$ 352. It is evident that under the same settings, the UDCNet method exhibits great competitiveness under FLOPs and parameters against some ORSIs-SOD methods ($e.g.$, ACCoNet \cite{ACCorNet}, EMFINet \cite{EMFINet}, and EPRNet \cite{ERPNet}). This is because the complexity of the FFT and the IFFT is only O($n$log$n$), while the large-kernel convolution in the designed component adopts depthwise separable convolution, with a complexity of O($n$).

\begin{table}[]
\renewcommand{\arraystretch}{1}
	\setlength{\tabcolsep}{1pt}
	\centering
	\caption{Comparison results of our FSDT block compared to Restormer \cite{Restormer} and Swin Transformer \cite{Swin}.}
	\resizebox{0.48\textwidth}{7mm}{
\begin{tabular}{c|c|c|cccccc|cccccc}
\hline\hline
\multirow{2}{*}{Method} & \multirow{2}{*}{\begin{tabular}[c]{@{}c@{}}Parameters\\ (M)\end{tabular}} & \multirow{2}{*}{\begin{tabular}[c]{@{}c@{}}FLOPs\\ (G)\end{tabular}} & \multicolumn{6}{c|}{ORSSD (200 images)}             & \multicolumn{6}{c}{EORSSD (600 images)}             \\
                        &                                                                           &                                                                       &\cellcolor{blue!15}$\mathcal{M}$$\downarrow$   & \cellcolor{blue!15}$F_{\varphi}^{m}$$\uparrow$   & \cellcolor{blue!15}$F_{\varphi}^{a}$$\uparrow$   & \cellcolor{blue!15}$F_{\varphi}^{w}$$\uparrow$   & \cellcolor{blue!15}$S_m$$\uparrow$     & \cellcolor{blue!15}$E_m$$\uparrow$     & \cellcolor{blue!15}$\mathcal{M}$$\downarrow$   & \cellcolor{blue!15}$F_{\varphi}^{m}$$\uparrow$   & \cellcolor{blue!15}$F_{\varphi}^{a}$$\uparrow$   & \cellcolor{blue!15}$F_{\varphi}^{w}$$\uparrow$   & \cellcolor{blue!15}$S_m$$\uparrow$     & \cellcolor{blue!15}$E_m$$\uparrow$    \\ \hline
Restormer  \cite{Restormer}                & \textbf{\color{red}41.85}                                                                     & \textbf{\color{red}54.93}                                                                & 0.0102 & 0.9122 & 0.8538 & 0.8760 & 0.9192 & 0.9632 & 0.0088 & 0.8706 &  \textbf{\color{red}0.8116} & 0.8411 & 0.8818 &  \textbf{\color{red}0.9501} \\ 
Swin.  \cite{Swin}                    & 46.35                                                                     & 78.31                                                                & 0.0112 & 0.9093 &  \textbf{\color{red}0.8619} & 0.8789 & 0.9196 & 0.9697 & 0.0083 & 0.8768 & 0.7947 & 0.8410 &  \textbf{\color{red}0.8857} & 0.9429 \\ \hline
FSDT                 & 43.40                                                                      & 61.47                                                                &  \textbf{\color{red}0.0078} &  \textbf{\color{red}0.9148} & 0.8605 &  \textbf{\color{red}0.8882} &  \textbf{\color{red}0.9258} &  \textbf{\color{red}0.9710} &  \textbf{\color{red}0.0061} &  \textbf{\color{red}0.8844} & 0.7937 &  \textbf{\color{red}0.8479} & 0.8814 & 0.9460 \\ \hline\hline
\end{tabular}}
\label{FSDT-S-R}
\end{table}

\begin{figure}[t]
\centering\includegraphics[width=0.48\textwidth,height=4cm]{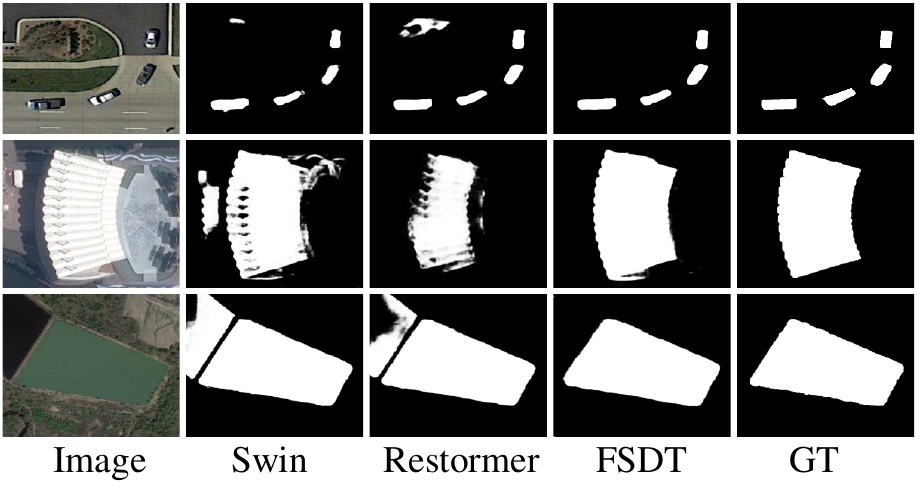}
	\captionsetup{font={small}, justification=raggedright}
	\caption{Visual predicted results of the proposed FSDT block and other Transformer blocks ($i.e.$, Restormer \cite{Restormer} and Swin \cite{Swin}).}
	\label{Visual_FSDT_R_S}
\end{figure}

\begin{table}[t]
\renewcommand{\arraystretch}{1}
	\setlength{\tabcolsep}{3pt}
	\centering
	\caption{Ablation analysis of our DJO decoder structure.}
	\resizebox{0.48\textwidth}{7mm}{
\begin{tabular}{c|cccccc|cccccc}
\hline\hline
\multirow{2}{*}{Methods} & \multicolumn{6}{c|}{ORSSD (200 images)}             & \multicolumn{6}{c}{EORSSD (600 images)}             \\
                          &\cellcolor{blue!15}$\mathcal{M}$$\downarrow$   & \cellcolor{blue!15}$F_{\varphi}^{m}$$\uparrow$   & \cellcolor{blue!15}$F_{\varphi}^{a}$$\uparrow$   & \cellcolor{blue!15}$F_{\varphi}^{w}$$\uparrow$   & \cellcolor{blue!15}$S_m$$\uparrow$     & \cellcolor{blue!15}$E_m$$\uparrow$     & \cellcolor{blue!15}$\mathcal{M}$$\downarrow$   & \cellcolor{blue!15}$F_{\varphi}^{m}$$\uparrow$   & \cellcolor{blue!15}$F_{\varphi}^{a}$$\uparrow$   & \cellcolor{blue!15}$F_{\varphi}^{w}$$\uparrow$   & \cellcolor{blue!15}$S_m$$\uparrow$     & \cellcolor{blue!15}$E_m$$\uparrow$     \\ \hline
DJO W/o edge             & 0.0089 & 0.9183 & 0.8711 & 0.8931 & \textbf{\color{red}0.9303} & 0.9706 & 0.0071 & 0.8693 & 0.7575 & 0.8194 & 0.8772 & 0.9209 \\ \hline
DJO                      & \textbf{\color{red}0.0070} & \textbf{\color{red}0.9212} & \textbf{\color{red}0.8827} & \textbf{\color{red}0.9013} & 0.9297 & \textbf{\color{red}0.9760} & \textbf{\color{red}0.0066} & \textbf{\color{red}0.8697} & \textbf{\color{red}0.7769} & \textbf{\color{red}0.8306} & \textbf{\color{red}0.8830} & \textbf{\color{red}0.9326} \\ \hline\hline
\end{tabular}}
\label{table5}
\end{table}

\begin{figure}[t]	
        \centering\includegraphics[width=0.45\textwidth,height=3.2cm]{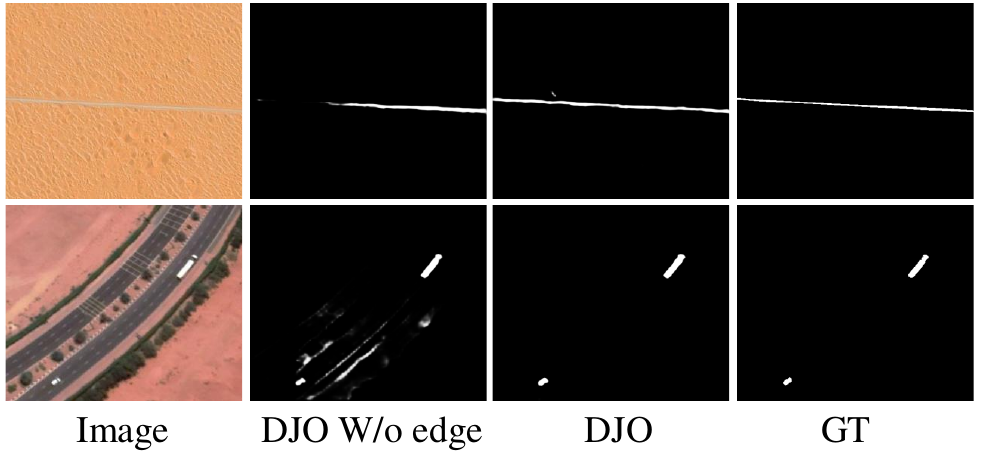}
	\captionsetup{font={small}, justification=raggedright}
	\caption{ Visual comparisons of the proposed DJO decoder. }
	\label{Fig.11}

\end{figure}

\begin{table}[t]
\renewcommand{\arraystretch}{1}
	\setlength{\tabcolsep}{2pt}
	\centering
	\caption{Comparison results of the proposed DSE module compared to ASPP \cite{ASPP} and RFB \cite{RFB}.}
	\resizebox{0.48\textwidth}{8mm}{
\begin{tabular}{c|cccccc|cccccc}
\hline\hline
\multirow{2}{*}{Method} & \multicolumn{6}{c|}{ORSSD (200 images)}             & \multicolumn{6}{c}{EORSSD (600 images)}             \\
                         &\cellcolor{blue!15}$\mathcal{M}$$\downarrow$   & \cellcolor{blue!15}$F_{\varphi}^{m}$$\uparrow$   & \cellcolor{blue!15}$F_{\varphi}^{a}$$\uparrow$   & \cellcolor{blue!15}$F_{\varphi}^{w}$$\uparrow$   & \cellcolor{blue!15}$S_m$$\uparrow$     & \cellcolor{blue!15}$E_m$$\uparrow$     & \cellcolor{blue!15}$\mathcal{M}$$\downarrow$   & \cellcolor{blue!15}$F_{\varphi}^{m}$$\uparrow$   & \cellcolor{blue!15}$F_{\varphi}^{a}$$\uparrow$   & \cellcolor{blue!15}$F_{\varphi}^{w}$$\uparrow$   & \cellcolor{blue!15}$S_m$$\uparrow$     & \cellcolor{blue!15}$E_m$$\uparrow$     \\ \hline
ASPP \cite{ASPP}                    & \textbf{\color{red}0.0127} & 0.8816 & 0.7981 & 0.8328 & 0.8900 & 0.9358 & 0.0096 & 0.8383 & 0.7295 & 0.7881 & 0.8536 & 0.9097 \\ 
RFB \cite{RFB}                     & 0.0132 & 0.8703 & 0.7901 & 0.8234 & 0.8862 & 0.9330 & \textbf{\color{red}0.0077} & 0.8752 & 0.7772 & 0.8296 & 0.8700 & 0.9326 \\ \hline
DSE                  & \textbf{\color{red}0.0127} & \textbf{\color{red}0.8907} & \textbf{\color{red}0.8081} & \textbf{\color{red}0.8405} & \textbf{\color{red}0.8953} & \textbf{\color{red}0.9394} & 0.0080 & \textbf{\color{red}0.8767} & \textbf{\color{red}0.7926} & \textbf{\color{red}0.8390} & \textbf{\color{red}0.8803} & \textbf{\color{red}0.9407} \\ \hline\hline
\end{tabular}}
\label{table6}
\end{table}

\begin{figure}[t]	
        \centering\includegraphics[width=0.46\textwidth,height=4cm]{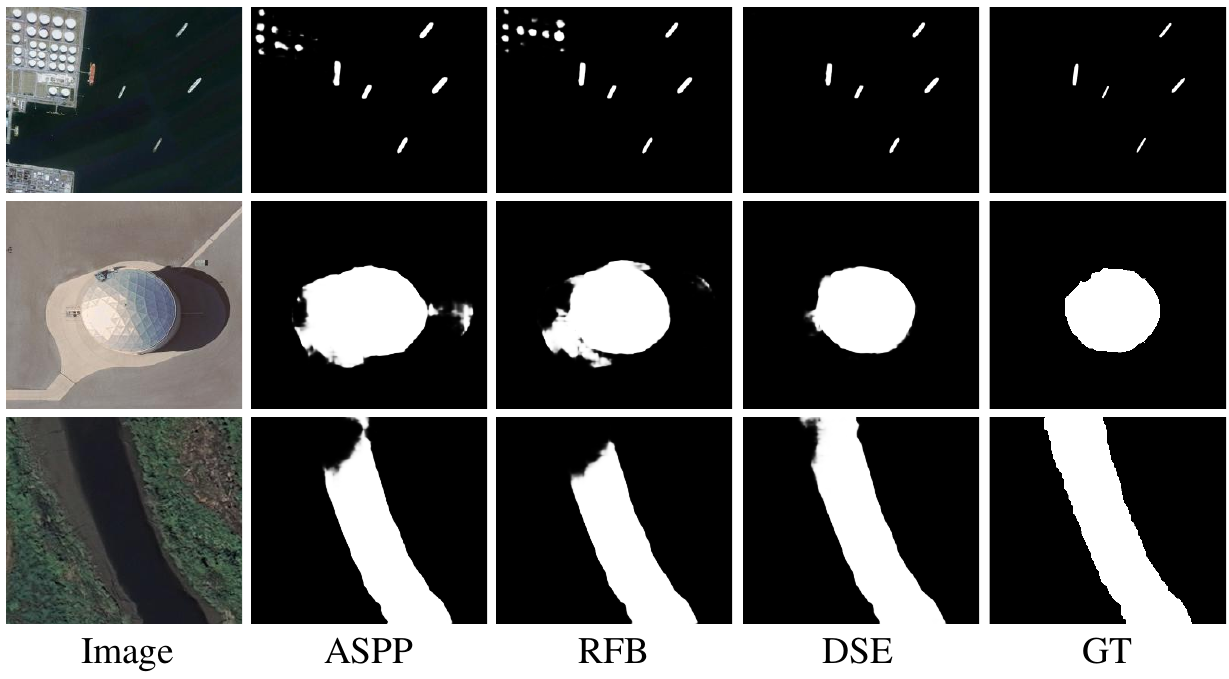}
	\captionsetup{font={small}, justification=raggedright}
	\caption{ Visual predicted results of the proposed DSE module and other enhancement modules ($i.e.$, ASPP \cite{ASPP} and RFB \cite{RFB}). }
	\label{Fig.12}

\end{figure}

\begin{table}[t]
\renewcommand{\arraystretch}{1}
	\setlength{\tabcolsep}{2pt}
	\centering
	\caption{Dilation rate analysis in the DSE module.}
	\resizebox{0.48\textwidth}{10mm}{
\begin{tabular}{c|c|cccccc|cccccc}
\hline\hline
\multirow{2}{*}{Num.} & \multirow{2}{*}{Settings} & \multicolumn{6}{c|}{ORSSD(200 images)}              & \multicolumn{6}{c}{EORSSD(600 images)}              \\
                      &                           &\cellcolor{blue!15}$\mathcal{M}$$\downarrow$   & \cellcolor{blue!15}$F_{\varphi}^{m}$$\uparrow$   & \cellcolor{blue!15}$F_{\varphi}^{a}$$\uparrow$   & \cellcolor{blue!15}$F_{\varphi}^{w}$$\uparrow$   & \cellcolor{blue!15}$S_m$$\uparrow$     & \cellcolor{blue!15}$E_m$$\uparrow$     & \cellcolor{blue!15}$\mathcal{M}$$\downarrow$   & \cellcolor{blue!15}$F_{\varphi}^{m}$$\uparrow$   & \cellcolor{blue!15}$F_{\varphi}^{a}$$\uparrow$   & \cellcolor{blue!15}$F_{\varphi}^{w}$$\uparrow$   & \cellcolor{blue!15}$S_m$$\uparrow$     & \cellcolor{blue!15}$E_m$$\uparrow$     \\ \hline
(a)                   & (2, 3, 4, 5)                 & 0.0142 & 0.8758 & 0.8013 & 0.8369 & 0.8802 & 0.9399 & 0.0091 & 0.8511 & 0.7384 & 0.8013 & 0.8613 & 0.9157 \\ 
(b)                   & (3, 6, 9, 12)                & 0.0147 & 0.8736 & 0.7932 & 0.8285 & 0.8806 & 0.9340 & 0.0092 & 0.8201 & 0.7068 & 0.7553 & 0.8546 & 0.8949 \\ 
(c)                   & (2, 4, 8, 12)                & 0.0147 & 0.8763 & 0.8060 & 0.8349 & 0.8892 & \textbf{\color{red}0.9395} & 0.0100 & 0.8272 & 0.7032 & 0.7621 & 0.8435 & 0.8932 \\ 
(d)                   & (5, 7, 9, 11)                & 0.0136 & 0.8752 & 0.7965 & 0.8298 & 0.8910 & 0.9354 & 0.0095 & 0.8108 & 0.7137 & 0.7561 & 0.8544 & 0.9024 \\ \hline
(e)                   & (3, 6, 12, 18)               & \textbf{\color{red}0.0127} & \textbf{\color{red}0.8907} & \textbf{\color{red}0.8081} & \textbf{\color{red}0.8405} & \textbf{\color{red}0.8953} & 0.9394 & \textbf{\color{red}0.0080} & \textbf{\color{red}0.8767} & \textbf{\color{red}0.7926} & \textbf{\color{red}0.8390} & \textbf{\color{red}0.8803} & \textbf{\color{red}0.9407} \\ \hline\hline
\end{tabular}}

\label{table7}
\end{table}

\subsection{Ablation Study}
In this section, we present a comprehensive analysis of the effectiveness and rationality of each component ($i.e.$, FSDT block, DJO decoder, and DSE module) in our UDCNet method. We conduct extensive ablation studies using both the ORSSD \cite{ORSSD} and EORSSD \cite{EORSSD} datasets, with all methods being implemented based on the ResNet50 \cite{ResNet} network. Furthermore, we investigate experimental analysis on the hyper-parameter of the UDCNet model to evaluate its performance.

\subsubsection{\textbf{Effect of FSDT block}}
FSDT block comprises four essential components, including FPSA, SPSA, AFS, and CDFFN. These components are meticulously designed to adaptively model long-range dependencies in initial features from the frequency and spatial domains, enhancing the inference of remote sensing objects. As shown in Table \ref{table3}, the initial ``Baseline'' (Tab. \ref{table3}(a)) consists of ResNet50 \cite{ResNet} and a feature pyramid network (FPN) \cite{FPN}. Then we integrate the ``FSDT'' block (Tab. \ref{table3}(b)) in the ``Baseline''. It is evident that on the ORSSD \cite{ORSSD} dataset, there is a significant improvement in performance across six evaluation metrics. The performance changes are as follows: from 0.0143 to 0.0078, from 0.8641 to 0.9148, from 0.7780 to 0.8605, from 0.8144 to 0.8882, from 0.8751 to 0.9258, and from 0.9218 to 0.9710. Fig. \ref{Fig.8} presents visual results of different components, and it can be observed that the embedding of the FSDT block is beneficial for improving the detection accuracy of the baseline. Meanwhile, we analyze and verify the internal structure of the FSDT block, as shown in Table \ref{table4} and Fig .\ref{Fig.9}. ``FSDT W/Fre" refers to the FSDT block that only contains FPSA and frequency information feed-forward network. ``FSDT W/Spa" denotes the FSDT block that only contains spatial information, while ``FSDT W/o AFS" indicates that the frequency and spatial information are not fused using adaptive fusion strategy (AFS) in the FSDT block. It can be seen that the performance of ``FSDT W/Fre" or ``FSDT W/Spa" alone is significantly inferior to that of ``FSDT W/o AFS", and the performance of ``FSDT W/o AFS" is further improved after adaptive fusion. Furthermore, we conduct a performance comparison with some existing traditional self-attention-based architectures ($e.g.$, Restormer \cite{Restormer} and Swin Transformer \cite{Swin}). As shown in Table \ref{FSDT-S-R} and Fig. \ref{Visual_FSDT_R_S}, under a similar number of parameters, our proposed FSDT block is highly competitive. This is attributed to the collaboration between the frequency and spatial domains. These quantitative and qualitative results demonstrate the rationality of the structure of the FSDT block and prove that there is complementarity between frequency and spatial information, which is beneficial for detecting remote sensing objects.

\begin{table}[t]
\renewcommand{\arraystretch}{1}
	\setlength{\tabcolsep}{1pt}
	\centering
	\caption{Hyper-parameter analysis of our UDCNet method.}
	\resizebox{0.48\textwidth}{9mm}{
\begin{tabular}{c|c|c|c|cccccc|cccccc}
\hline\hline
\multirow{2}{*}{Num.} & Settings        & \multirow{2}{*}{\begin{tabular}[c]{@{}c@{}}FLOPs\\ (G)\end{tabular}} & \multirow{2}{*}{\begin{tabular}[c]{@{}c@{}}Parameters\\ (M)\end{tabular}} & \multicolumn{6}{c|}{ORSSD (200 images)}             & \multicolumn{6}{c}{EORSSD (600 images)}             \\ \cline{2-2}
                      & Decoder channel &                                                                      &                                                                            &\cellcolor{blue!15}$\mathcal{M}$$\downarrow$   & \cellcolor{blue!15}$F_{\varphi}^{m}$$\uparrow$   & \cellcolor{blue!15}$F_{\varphi}^{a}$$\uparrow$   & \cellcolor{blue!15}$F_{\varphi}^{w}$$\uparrow$   & \cellcolor{blue!15}$S_m$$\uparrow$     & \cellcolor{blue!15}$E_m$$\uparrow$     & \cellcolor{blue!15}$\mathcal{M}$$\downarrow$   & \cellcolor{blue!15}$F_{\varphi}^{m}$$\uparrow$   & \cellcolor{blue!15}$F_{\varphi}^{a}$$\uparrow$   & \cellcolor{blue!15}$F_{\varphi}^{w}$$\uparrow$   & \cellcolor{blue!15}$S_m$$\uparrow$     & \cellcolor{blue!15}$E_m$$\uparrow$     \\ \hline
(a)                   & 64              & \textbf{\color{red}35.00}                                                                & \textbf{\color{red}49.04}                                                                     & 0.0077 & 0.9223 & 0.8886 & 0.9015 & 0.9336 & 0.9763 & 0.0060 & 0.8850 & 0.8252 & 0.8586 & 0.8931 & 0.9579 \\ 
(b)                   & 96              & 62.67                                                                & 58.79                                                                     & 0.0077 & 0.9236 & 0.8919 & 0.9058 & 0.9381 & 0.9784 & 0.0059 & 0.8827 & 0.8085 & 0.8535 & 0.8927 & 0.9539 \\ 
(c)                   & 128             & 101.19                                                               & 72.20                                                                     & \textbf{\color{red}0.0068} & \textbf{\color{red}0.9267} & \textbf{\color{red}0.8932} & 0.9093 & \textbf{\color{red}0.9389} & 0.9770 & \textbf{\color{red}0.0056} & 0.8903 & 0.8211 & 0.8615 & 0.8957 & 0.9547 \\ 
(d)                   & 160             & 150.56                                                               & 89.27                                                                     & 0.0071 & 0.9256 & {0.8928} & \textbf{\color{red}0.9107} & 0.9377 & \textbf{\color{red}0.9800} & 0.0058 & \textbf{\color{red}0.8929} & \textbf{\color{red}0.8313} & \textbf{\color{red}0.8697} & \textbf{\color{red}0.8982} & \textbf{\color{red}0.9626} \\ 
(e)                   & 256             & 363.73                                                               & 162.46                                                                    & 0.0070 & 0.9252 & 0.8910 & 0.9075 & 0.9365 & 0.9790 & 0.0059 & 0.8892 & 0.8306 & 0.8639 & 0.8956 & 0.9584 \\ \hline\hline
\end{tabular}}

\label{table8}
\end{table}

\subsubsection{\textbf{Effect of DJO decoder}}
DJO decoder aims to interact diverse information from different level features and strengthen the edge structure of salient objects. From Table \ref{table3}, ``DJO'' decoder (Tab. \ref{table3}(c)) exhibits significant performance improvements compared to FPN decoder in the ``Baseline'' (Tab. \ref{table3}(a)). Specifically, it achieves a substantial increase of 6.24\%, 8.44\% and 5.88\%, 9.82\% in terms of $S_m$ and $E_m$ on the ORSSD  \cite{ORSSD} and EORSSD \cite{EORSSD} datasets, respectively. Other evaluation metrics also obtain noticeable promotion. Furthermore, we conduct compatibility validation between the ``FSDT'' block and ``DJO'' decoder. As shown in Table \ref{table3}(e), it is evident that distinct advantages over using ``FSDT'' block (Tab. \ref{table3}(b)) or ``DJO'' decoder (Tab. \ref{table3}(c)) individually. Furthermore, we have validated the effectiveness of the edge enhancement branch. In Table \ref{table5} and Fig .\ref{Fig.11}, upon removing the edge branch from the DJO decoder, a noticeable decline in model performance is observed. This demonstrates the edge enhancement branch's capability to refine the edge contours.

\subsubsection{\textbf{Effect of DSE module}}
The purpose of the DSE module is to obtain higher-level semantic information from different receptive fields, using well-designed atrous convolutions to assist in the perception of remote sensing objects. From Table \ref{table3} (d), (f), and (g), it can be observed that through the semantic supplementation of the ``DSE'' module, there is a significant increase in the detection accuracy of the ``Baseline'', ``FSDT'' block, and ``DJO'' decoder. Moreover, as illustrated in Table \ref{table6} and Fig. \ref{Fig.12}, we have conducted a comparative analysis between our DSE module and existing modules such as ``ASPP'' \cite{ASPP} and ``RFB'' \cite{RFB}. The results demonstrate the significant superiority of our DSE module over the others. Additionally, we analyze the setting of the dilation rates from atrous convolutions within the DSE module in Table \ref{table7}. We test several different designs and observe that the designed DSE module performs best when set to a specific value (3,6,12,18). These above results provide clear evidence for the effectiveness and rationality of the proposed DSE module.

\subsubsection{\textbf{Hyper-parameter analysis}}
To further validate the effectiveness of the model, we conduct experimental analysis on the hyper-parameters ($i.e.$, channels in the decoder) of the model. As presented in Table \ref{table8}, we test different channels in the decoder, including 64 (Tab. \ref{table8}(a)), 96 (Tab. \ref{table8}(b)), 128 (Tab. \ref{table8}(c)), 160 (Tab. \ref{table8}(d)), and 256 (Tab. \ref{table8}(e)). It can be observed that the proposed UDCNet method achieves superior performance across different channel settings. Although increasing the channels may improve the model's performance to some extent, it significantly increases the parameters and FLOPs, which slows down the inference speed and makes the model more difficult to converge. Therefore, considering all factors, we set the channel hyper-parameter to 128.
\begin{figure}[h]
	\centering\includegraphics[width=0.46\textwidth,height=2.5cm]{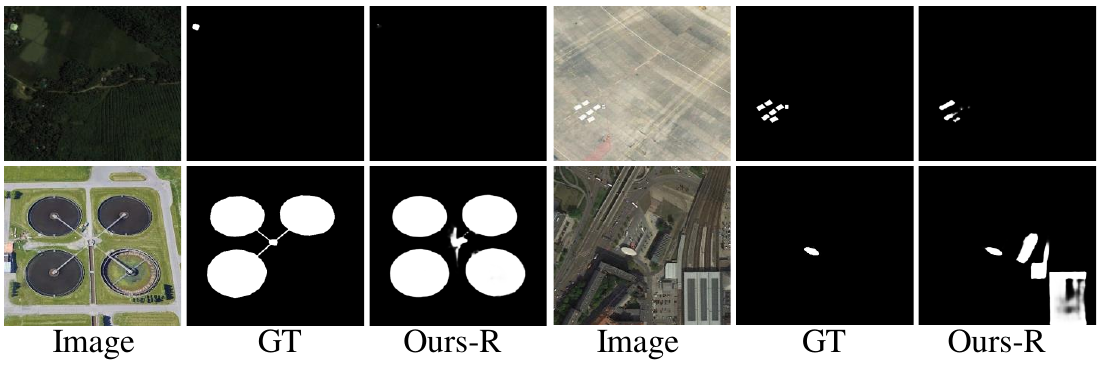}
	\captionsetup{font={small}, justification=raggedright}
	\caption{Failed prediction results.``GT'' denotes the ground truth.}
	\label{Fig.13}

\end{figure}
\subsection{Failure Cases}
In Fig. \ref{Fig.13}, some failure cases of our UDCNet model are presented, indicating its limitations in challenging scenarios. As shown in the first row in Fig. \ref{Fig.13}, when the remote sensing objects are extremely small or numerous, the proposed UDCNet method can only detect some of them and fail to segment them completely. The second row of images demonstrates that our UDCNet method may make misjudgments on remote sensing objects when there is a certain similarity between the object and background semantics in the scene. 

In future work, we alleviate these issues by considering two aspects. Firstly, we will adopt an amplification strategy to segment extremely small targets. Secondly, we will introduce larger models, such as SAM \cite{SAM}, to enhance the semantic understanding of image content.
\begin{table*}[t]
\renewcommand{\arraystretch}{1}
	\setlength{\tabcolsep}{2pt}
	\centering
	\caption{ Quantitative results on four widely-used SOD datasets. The best three results are shown in {\color{red} red}, {\color{green} green}, and {\color{blue} blue}. ``$\uparrow$/$\downarrow$'' present that a higher/lower value is better. ``Ours-R'' and  ``Ours-P'' denote ResNet50 \cite{ResNet} or PVT \cite{Pvt2} as backbone.}
	\resizebox*{0.90\textwidth}{66mm}{
\begin{tabular}{ccccccccccccccccccccccccc}
\hline\hline
\multicolumn{1}{c|}{\multirow{2}{*}{\textbf{Methods}}} & \multicolumn{6}{c|}{\textbf{PASCAL-S (850 images)}}                          & \multicolumn{6}{c|}{\textbf{ECSSD (1000 images)}}                            & \multicolumn{6}{c|}{\textbf{HKU-IS (4447 images)}}                           & \multicolumn{6}{c}{\textbf{DUTS-TE (5019 images)}}      \\
\multicolumn{1}{c|}{}                        &\cellcolor{blue!15}$\mathcal{M}$$\downarrow$   &\cellcolor{blue!15} $F_{\varphi}^{m}$$\uparrow$   & \cellcolor{blue!15}$F_{\varphi}^{a}$$\uparrow$   &\cellcolor{blue!15} $F_{\varphi}^{w}$$\uparrow$   &\cellcolor{blue!15} $S_m$$\uparrow$     & \multicolumn{1}{c|}{\cellcolor{blue!15}$E_m$$\uparrow$}    & \cellcolor{blue!15}$\mathcal{M}$$\downarrow$   &\cellcolor{blue!15} $F_{\varphi}^{m}$$\uparrow$   & \cellcolor{blue!15}$F_{\varphi}^{a}$$\uparrow$   & \cellcolor{blue!15}$F_{\varphi}^{w}$$\uparrow$   & \cellcolor{blue!15}$S_m$$\uparrow$     & \multicolumn{1}{c|}{\cellcolor{blue!15}$E_m$$\uparrow$}    & \cellcolor{blue!15}$\mathcal{M}$$\downarrow$   & \cellcolor{blue!15}$F_{\varphi}^{m}$$\uparrow$   & \cellcolor{blue!15}$F_{\varphi}^{a}$$\uparrow$   & \cellcolor{blue!15}$F_{\varphi}^{w}$$\uparrow$   & \cellcolor{blue!15}$S_m$$\uparrow$     & \multicolumn{1}{c|}{\cellcolor{blue!15}$E_m$$\uparrow$}    & \cellcolor{blue!15}$\mathcal{M}$$\downarrow$   & \cellcolor{blue!15}$F_{\varphi}^{m}$$\uparrow$   & \cellcolor{blue!15}$F_{\varphi}^{a}$$\uparrow$   & \cellcolor{blue!15}$F_{\varphi}^{w}$$\uparrow$   & \cellcolor{blue!15}$S_m$$\uparrow$     & \multicolumn{1}{c|}{\cellcolor{blue!15}$E_m$$\uparrow$}    \\ \hline
\multicolumn{25}{c}{Convolutional Neural Network-based SOD Methods in Natural Scene Images.}                                                                                                                                                         \\ \hline
\multicolumn{1}{c|}{PoolNet \cite{PoolNet}}                 & 0.075 & 0.863 & 0.815 & 0.793 & 0.849 & \multicolumn{1}{c|}{0.876} & 0.039 & 0.944 & 0.915 & 0.896 & 0.921 & \multicolumn{1}{c|}{0.945} & 0.032 & 0.934 & 0.900 & 0.883 & 0.915 & \multicolumn{1}{c|}{0.955} & 0.040 & 0.880 & 0.809 & 0.807 & 0.882 & 0.904 \\ 
\multicolumn{1}{c|}{CPD \cite{CPD}}                     & 0.071 & 0.860 & 0.820 & 0.794 & 0.848 & \multicolumn{1}{c|}{0.887} & 0.037 & 0.939 & 0.917 & 0.898 & 0.918 & \multicolumn{1}{c|}{0.950} & 0.033 & 0.927 & 0.895 & 0.879 & 0.908 & \multicolumn{1}{c|}{0.952} & 0.043 & 0.866 & 0.805 & 0.795 & 0.868 & 0.904 \\ 
\multicolumn{1}{c|}{ITSD \cite{ITSD}}                    & 0.066 & 0.870 & 0.785 & 0.812 & 0.859 & \multicolumn{1}{c|}{0.863} & 0.035 & 0.947 & 0.895 & 0.911 & 0.925 & \multicolumn{1}{c|}{0.932} & 0.031 & 0.934 & 0.899 & 0.894 & 0.917 & \multicolumn{1}{c|}{0.953} & 0.041 & 0.883 & 0.804 & 0.824 & 0.884 & 0.898 \\ 
\multicolumn{1}{c|}{GateNet \cite{GateNet}}                 & 0.067 & 0.869 & 0.819 & 0.797 & 0.858 & \multicolumn{1}{c|}{0.884} & 0.040 & 0.945 & 0.916 & 0.894 & 0.920 & \multicolumn{1}{c|}{0.943} & 0.033 & 0.934 & 0.899 & 0.880 & 0.915 & \multicolumn{1}{c|}{0.953} & 0.040 & 0.888 & 0.807 & 0.809 & 0.884 & 0.903 \\ 
\multicolumn{1}{c|}{RASNet \cite{RASNet}}                  & 0.068 & 0.862 & 0.820 & 0.804 & 0.853 & \multicolumn{1}{c|}{0.883} & 0.036 & 0.944 & 0.916 & 0.905 & 0.920 & \multicolumn{1}{c|}{0.941} & 0.030 & 0.932 & 0.906 & 0.892 & 0.914 & \multicolumn{1}{c|}{0.956} & 0.039 & 0.880 & 0.825 & 0.822 & 0.879 & 0.913 \\ 
\multicolumn{1}{c|}{MINet \cite{MINet}}                   & 0.064 & 0.867 & 0.829 & 0.809 & 0.856 & \multicolumn{1}{c|}{0.898} & 0.034 & 0.948 & 0.924 & 0.911 & 0.925 & \multicolumn{1}{c|}{0.953} & 0.029 & 0.935 & 0.909 & 0.897 & 0.918 & \multicolumn{1}{c|}{\color{blue}0.960} & \color{blue}0.037 & 0.884 & 0.828 & 0.825 & 0.883 & 0.917 \\ 
\multicolumn{1}{c|}{U2Net} \cite{U2Net}                   & 0.074 & 0.860 & 0.771 & 0.792 & 0.845 & \multicolumn{1}{c|}{0.850} & \color{blue}0.033 & \color{blue}0.951 & 0.891 & 0.910 & 0.927 & \multicolumn{1}{c|}{0.925} & 0.031 & 0.935 & 0.897 & 0.889 & 0.915 & \multicolumn{1}{c|}{0.949} & 0.045 & 0.872 & 0.791 & 0.803 & 0.872 & 0.887 \\ 
\multicolumn{1}{c|}{LDF \cite{LDF}}                     & \color{blue}0.060 & 0.874 & 0.843 & 0.822 & \color{blue}0.862 & \multicolumn{1}{c|}{\color{blue}0.905} & 0.034 & 0.950 & \color{blue}0.930 & 0.915 & 0.924 & \multicolumn{1}{c|}{0.951} & 0.028 & \color{blue}0.939 & \color{blue}0.914 & 0.904 & 0.919 & \multicolumn{1}{c|}{\color{blue}0.960} & \color{red}0.034 & 0.898 & \color{green}0.855 & \color{green}0.845 & \color{green}0.891 & \color{green}0.929 \\ 
\multicolumn{1}{c|}{CSF \cite{CSF}}                     & 0.069 & 0.874 & 0.823 & 0.807 & \color{blue}0.862 & \multicolumn{1}{c|}{0.884} & \color{blue}0.033 & \color{blue}0.951 & 0.925 & 0.911 & \color{green}0.930 & \multicolumn{1}{c|}{0.954} & -     & -     & -     & -     & -     & \multicolumn{1}{c|}{-}     & 0.038 & 0.890 & 0.823 & 0.823 & 0.889 & 0.914 \\ 
\multicolumn{1}{c|}{DCN \cite{DCN}}                     & 0.062 & 0.872 & 0.837 & 0.820 & 0.861 & \multicolumn{1}{c|}{0.902} & \color{green}0.032 & \color{green}0.952 & 0.934 & \color{green}0.920 & 0.928 & \multicolumn{1}{c|}{\color{green}0.958} & \color{blue}0.027 & \color{blue}0.939 & \color{blue}0.914 & 0.905 & \color{blue}0.921 & \multicolumn{1}{c|}{\color{green}0.962} & \color{green}0.035 & \color{blue}0.894 & 0.844 & \color{blue}0.840 & \color{blue}0.890 & 0.923 \\ 
\multicolumn{1}{c|}{DSPN \cite{DSPN}}                    & 0.067 & 0.874 & 0.819 & 0.801 & \color{green}0.864 & \multicolumn{1}{c|}{0.883} & 0.039 & 0.945 & 0.910 & 0.891 & 0.924 & \multicolumn{1}{c|}{0.942} & 0.035 & 0.932 & 0.893 & 0.873 & 0.915 & \multicolumn{1}{c|}{0.951} & 0.043 & 0.883 & 0.791 & 0.794 & 0.881 & 0.892 \\ 
\multicolumn{1}{c|}{EDN \cite{EDN}}                     & 0.062 & \color{green}0.880 & \color{blue}0.847 & \color{green}0.827 & \color{green}0.864 & \multicolumn{1}{c|}{0.904} & \color{green}0.032 & \color{blue}0.951 & \color{green}0.933 & \color{blue}0.918 & 0.927 & \multicolumn{1}{c|}{\color{blue}0.955} & \color{green}0.026 & \color{green}0.941 & \color{green}0.919 & \color{blue}0.908 & \color{green}0.923 & \multicolumn{1}{c|}{\color{green}0.962} & \color{green}0.035 & \color{green}0.895 & 0.852 & \color{green}0.845 & \color{green}0.891 & \color{blue}0.928 \\ 
\multicolumn{1}{c|}{DCENet \cite{DCENet}}                  & 0.061 & \color{green}0.880 & 0.845 & 0.825 & \color{blue}0.862 & \multicolumn{1}{c|}{0.903} & 0.035 & 0.947 & 0.926 & 0.913 & 0.921 & \multicolumn{1}{c|}{0.952} & 0.029 & 0.933 & 0.908 & 0.898 & 0.915 & \multicolumn{1}{c|}{0.957} & 0.038 & 0.884 & 0.842 & 0.834 & 0.881 & 0.922 \\ 
\multicolumn{1}{c|}{DNA \cite{DNA}}                     & 0.079 & 0.855 & 0.790 & 0.772 & 0.837 & \multicolumn{1}{c|}{0.861} & 0.042 & 0.940 & 0.891 & 0.883 & 0.915 & \multicolumn{1}{c|}{0.935} & 0.035 & 0.928 & 0.863 & 0.864 & 0.905 & \multicolumn{1}{c|}{0.937} & 0.046 & 0.873 & 0.747 & 0.765 & 0.859 & 0.864 \\ 
\multicolumn{1}{c|}{RCSBNet \cite{RCSBNet}}                 & \color{green}0.059 & \color{blue}0.876 & \color{green}0.848 & \color{blue}0.826 & 0.860 & \multicolumn{1}{c|}{\color{green}0.907} & 0.034 & 0.944 & 0.927 & 0.916 & 0.922 & \multicolumn{1}{c|}{0.948} & \color{blue}0.027 & 0.938 & \color{red}0.924 & \color{green}0.909 & 0.918 & \multicolumn{1}{c|}{0.959} & \color{green}0.035 & 0.889 & \color{red}0.856 & \color{blue}0.840 & 0.879 & 0.920 \\ 
\multicolumn{1}{c|}{ICON-R \cite{ICON}}                & 0.064 & \color{blue}0.876 & 0.833 & 0.818 & 0.861 & \multicolumn{1}{c|}{0.893} & \color{green}0.032 & 0.950 & 0.928 & \color{blue}0.918 & \color{blue}0.929 & \multicolumn{1}{c|}{0.954} & 0.029 & \color{blue}0.939 & 0.910 & 0.902 & 0.920 & \multicolumn{1}{c|}{0.959} & \color{blue}0.037 & 0.892 & 0.838 & 0.837 & 0.887 & 0.919 \\ \hline
\multicolumn{1}{c|}{\textbf{Ours-R}}                  & \color{red}0.054 & \color{red}0.886 & \color{red}0.854 & \color{red}0.842 & \color{red}0.873 & \multicolumn{1}{c|}{\color{red}0.918} & \color{red}0.025 & \color{red}0.955 & \color{red}0.936 & \color{red}0.931 & \color{red}\color{red}0.935 & \multicolumn{1}{c|}{\color{red}0.965} & \color{red}0.025 & \color{red}0.944 & \color{green}0.919 & \color{red}0.915 & \color{red}0.924 & \multicolumn{1}{c|}{\color{red}0.966} & \color{red}0.034 & \color{red}0.900 & \color{blue}0.853 & \color{red}0.853 & \color{red}0.893 & \color{red}0.930 \\ \hline
\multicolumn{25}{c}{Transformer-based SOD Methods in Natural Scene Images}                                                                                                                          \\ \hline
\multicolumn{1}{c|}{VST \cite{VST}}                     & 0.061 & 0.876 & 0.829 & 0.816 & 0.871 & \multicolumn{1}{c|}{0.902} & 0.033 & 0.951 & 0.920 & 0.910 & 0.932 & \multicolumn{1}{c|}{0.957} & 0.029 & 0.942 & 0.900 & 0.897 & 0.928 & \multicolumn{1}{c|}{0.960} & 0.037 & 0.890 & 0.818 & 0.828 & 0.895 & 0.916 \\ 
\multicolumn{1}{c|}{GLST \cite{GSLT}}                  & 0.052 & \color{green}0.897 & \color{blue}0.855 & 0.846 & \color{red}0.886 & \multicolumn{1}{c|}{\color{blue}0.920} & 0.025 & \color{blue}0.961 & 0.935 & 0.930 & \color{green}0.942 & \multicolumn{1}{c|}{\color{blue}0.965} & \color{blue}0.024 & 0.951 & 0.915 & 0.914 & \color{green}0.935 & \multicolumn{1}{c|}{0.967} & \color{blue}0.027 & \color{green}0.921 & 0.857 & 0.873 & \color{red}0.918 & 0.940 \\ 
\multicolumn{1}{c|}{ICON-P \cite{ICON}}                & \color{blue}0.051 & \color{blue}0.893 & 0.854 & \color{blue}0.847 & \color{green}0.882 & \multicolumn{1}{c|}{0.915} & \color{blue}0.024 & 0.959 & \color{blue}0.936 & \color{blue}0.933 & \color{blue}0.940 & \multicolumn{1}{c|}{0.962} & \color{green}0.022 & \color{green}0.952 & \color{blue}0.925 & \color{blue}0.924 & \color{blue}0.934 & \multicolumn{1}{c|}{\color{blue}0.970} & \color{green}0.026 & \color{red}0.923 & \color{blue}0.868 & \color{blue}0.882 & \color{green}0.916 & \color{blue}0.942 \\ 
\multicolumn{1}{c|}{BBRF \cite{BBRF}}                  & \color{red}0.049 & 0.891 & \color{red}0.869 & \color{green}0.856 & \color{blue}0.877 & \multicolumn{1}{c|}{\color{red}0.925} & \color{green}0.022 & \color{green}0.962 & \color{green}0.950 & \color{green}0.944 & 0.939 & \multicolumn{1}{c|}{\color{green}0.972} & \color{red}0.020 & \color{blue}0.951 & \color{red}0.937 & \color{green}0.932 & 0.932 & \multicolumn{1}{c|}{\color{green}0.972} & \color{red}0.025 & \color{blue}0.916 & \color{red}0.893 & \color{green}0.887 & \color{blue}0.907 & \color{red}0.953 \\ \hline
\multicolumn{1}{c|}{\textbf{Ours-P}}                  & \color{green}0.050 & \color{red}0.899 & \color{green}0.866 & \color{red}0.857 & \color{green}0.882 & \multicolumn{1}{c|}{\color{green}0.923} & \color{red}0.019 & \color{red}0.966 & \color{red}0.951 & \color{red}0.948 & \color{red}0.947 & \multicolumn{1}{c|}{\color{red}0.974} & \color{red}0.020 & \color{red}0.954 & \color{green}0.934 & \color{red}0.933 & \color{red}0.936 & \multicolumn{1}{c|}{\color{red}0.973} & \color{green}0.026 & \color{red}0.923 & \color{green}0.886 & \color{red}0.892 & \color{green}0.916 & \color{green}0.949 \\ \hline\hline
\end{tabular}}
\label{table9}

\end{table*}

\subsection{Expanded Applications}
We extend the proposed UDCNet method to the field of salient object detection (SOD) in natural scene images (NSIs) to further demonstrate its generalization ability and robustness. Fig. \ref{Fig.14} and Table \ref{table9} present qualitative and quantitative results of our UDCNet model on different NSIs-SOD datasets.
\subsubsection{Experimental details}
Similar to in ORSIs-SOD, the training parameter settings of our UDCNet model remain unchanged. The evaluated experiment is based on four public SOD datasets, including PASCAL-S \cite{PASCAL-S}, ECSSD \cite{ECSSD}, HKU-IS \cite{HKU}, and DUTS-TE \cite{DUTS}. We train our method under the DUTS-Train \cite{DUTS} dataset that contain 10,553 images. 	
\subsubsection{Result comparisons}
\begin{figure}[t]
	\centering\includegraphics[width=0.5\textwidth, height=4cm]{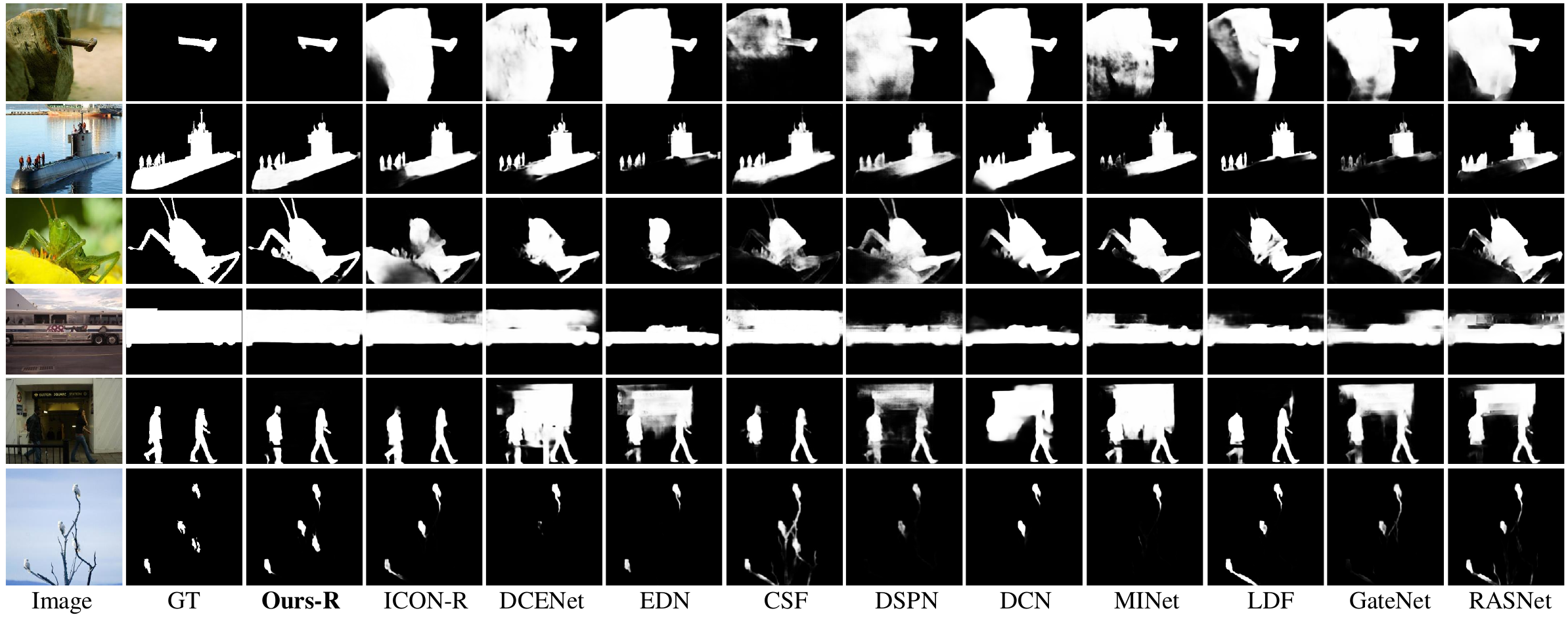}
	\captionsetup{font={small}, justification=raggedright}
	\caption{Visual predicted results of the proposed UDCNet method and 10 existing SOD methods.}
	\label{Fig.14}
\end{figure}

We compare the proposed UDCNet method with 20 SOTA NSIs-SOD methods, including PoolNet \cite{PoolNet}, CPD \cite{CPD}, ITSD \cite{ITSD}, GateNet \cite{GateNet}, RASNet \cite{RASNet}, U2Net \cite{U2Net}, MINet \cite{MINet}, LDF \cite{LDF}, CSF \cite{CSF}, DCN \cite{DCN}, DSPN \cite{DSPN}, EDN \cite{EDN}, DCENet \cite{DCENet}, DNA \cite{DNA}, RCSBNet \cite{RCSBNet}, ICON-R \cite{ICON}, VST \cite{VST}, GLST \cite{GSLT}, ICON-P \cite{ICON}, and BBRF \cite{BBRF}. From Table \ref{table9}, it can be seen that our UDCNet model has significant advantages compared to some current NSIs-SOD methods. As depicited in Fig. \ref{Fig.14}, we show the predicted saliency maps, it can be observed that our UDCNet method achieves great performance when facing salient objects of different scales and types. These above results clearly indicate that the proposed UDCNet method possesses a remarkable generalization ability. 

\section{Conclusions}
In this article, we propose a novel United Domain Cognition Network for accurate ORSIs-SOD tasks, called UDCNet. Our UDCNet model consists of three important components (namely, FSDT block, DSE module, and DJO decoder) designed for robust feature generation to alleviate the issue of the locality of features in the spatial domain. Specifically, in the FSDT block, initial features are divided into local spatial and global frequency features, which first adapt to weight adjustment through self-attention, followed by the interactive learning of optimized local and global features. Furthermore, the DSE module is constructed to obtain higher-level semantic for guiding object inference through atrous convolutions with different receptive fields. Finally, we utilize the DJO decoder to integrate multi-level features to generate accurate saliency maps through the collaboration of the saliency branch and edge branch. Extensive experiments have demonstrated the superiority and effectiveness of the proposed UDCNet method on three widely-used ORSIs-SOD datasets.

\ifCLASSOPTIONcaptionsoff
\newpage
\fi

\bibliographystyle{./IEEEtran}
\bibliography{./UDCNet}

\begin{IEEEbiography}
[{\includegraphics[width=1in,height=1.25in,clip,keepaspectratio]{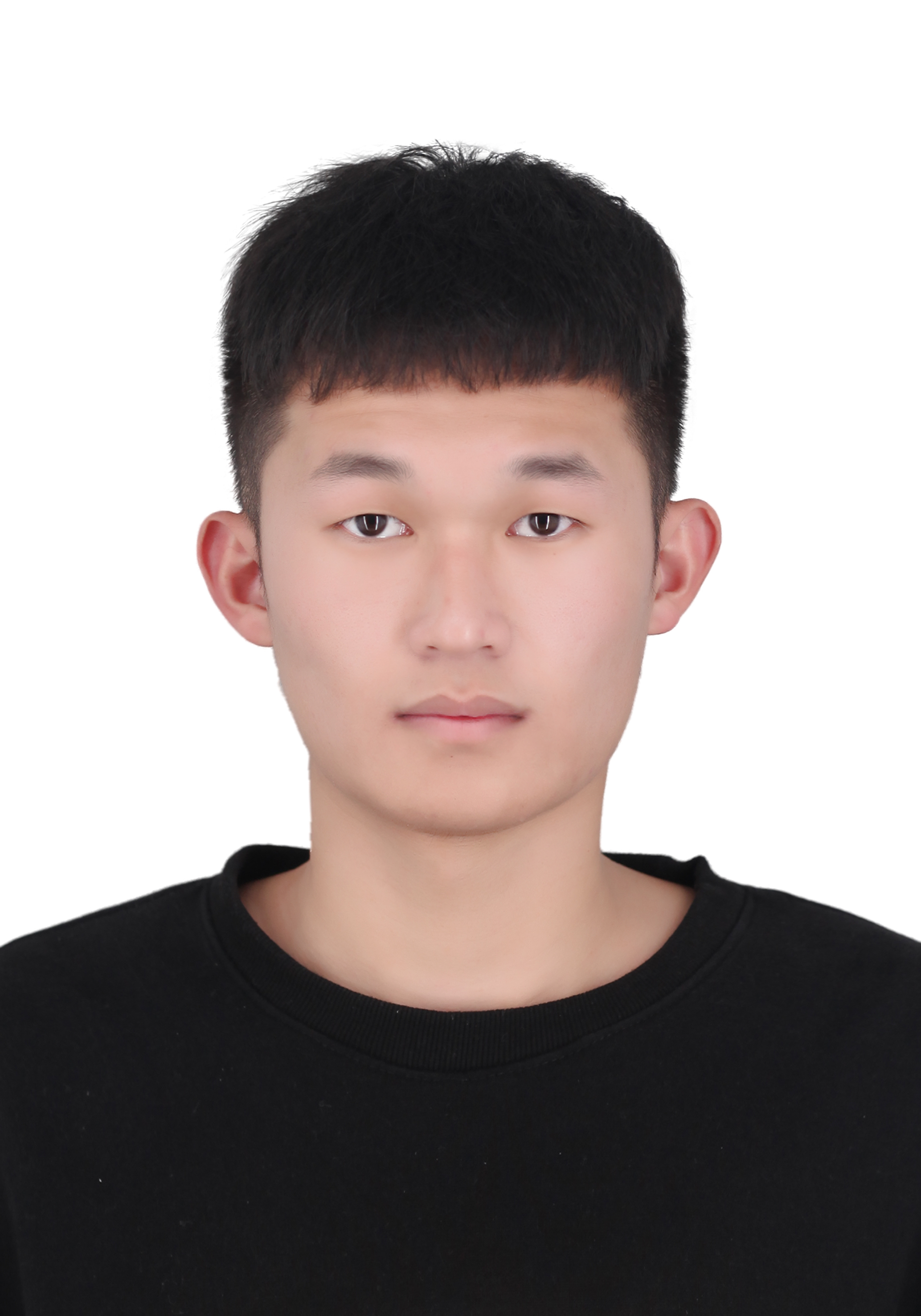}}]{Yanguang Sun} received the M.S. degree from Anhui University of Science and Technology (AUST), Huainan, Anhui, China, in 2023, majoring in software engineering. He is currently pursuing his Ph.D. at Nanjing University of Science and Technology (NJUST), Nanjing, Jiangsu, China, under the supervision of Professor Lei Luo. His research interests focus on computer vision and image analysis. He has served as a reviewer for prestigious conferences such as CVPR and AAAI.	
\end{IEEEbiography}

\begin{IEEEbiography}
[{\includegraphics[width=1in,height=1.25in,clip,keepaspectratio]{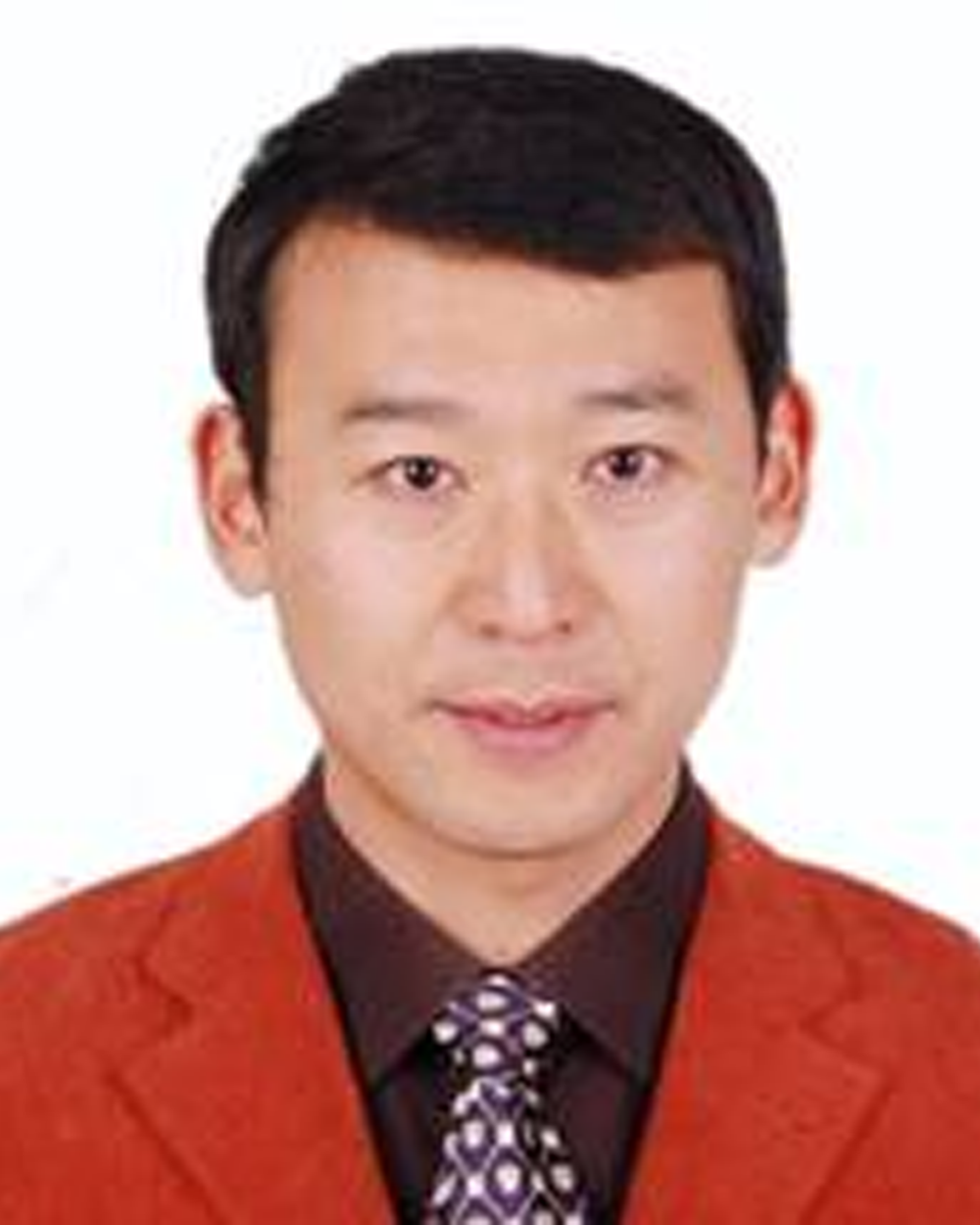}}]{Jian Yang} received the PhD degree from Nanjing University of Science and Technology (NJUST) in 2002, majoring in pattern recognition and intelligence systems. From 2003 to 2007, he was a Postdoctoral Fellow at the University of Zaragoza, Hong Kong Polytechnic University and New Jersey Institute of Technology, respectively. From 2007 to present, he is a professor in the School of Computer Science and Technology of NJUST. He is the author of more than 300 scientific papers in pattern recognition and computer vision. His papers have been cited over 49000 times in the Scholar Google. His research interests include pattern recognition and computer vision. Currently, he is/was an associate editor of Pattern Recognition, Pattern Recognition Letters, IEEE Trans. Neural Networks and Learning Systems, and Neurocomputing. He is a Fellow of IAPR.	
\end{IEEEbiography}

\begin{IEEEbiography}
[{\includegraphics[width=1in,height=1.25in,clip,keepaspectratio]{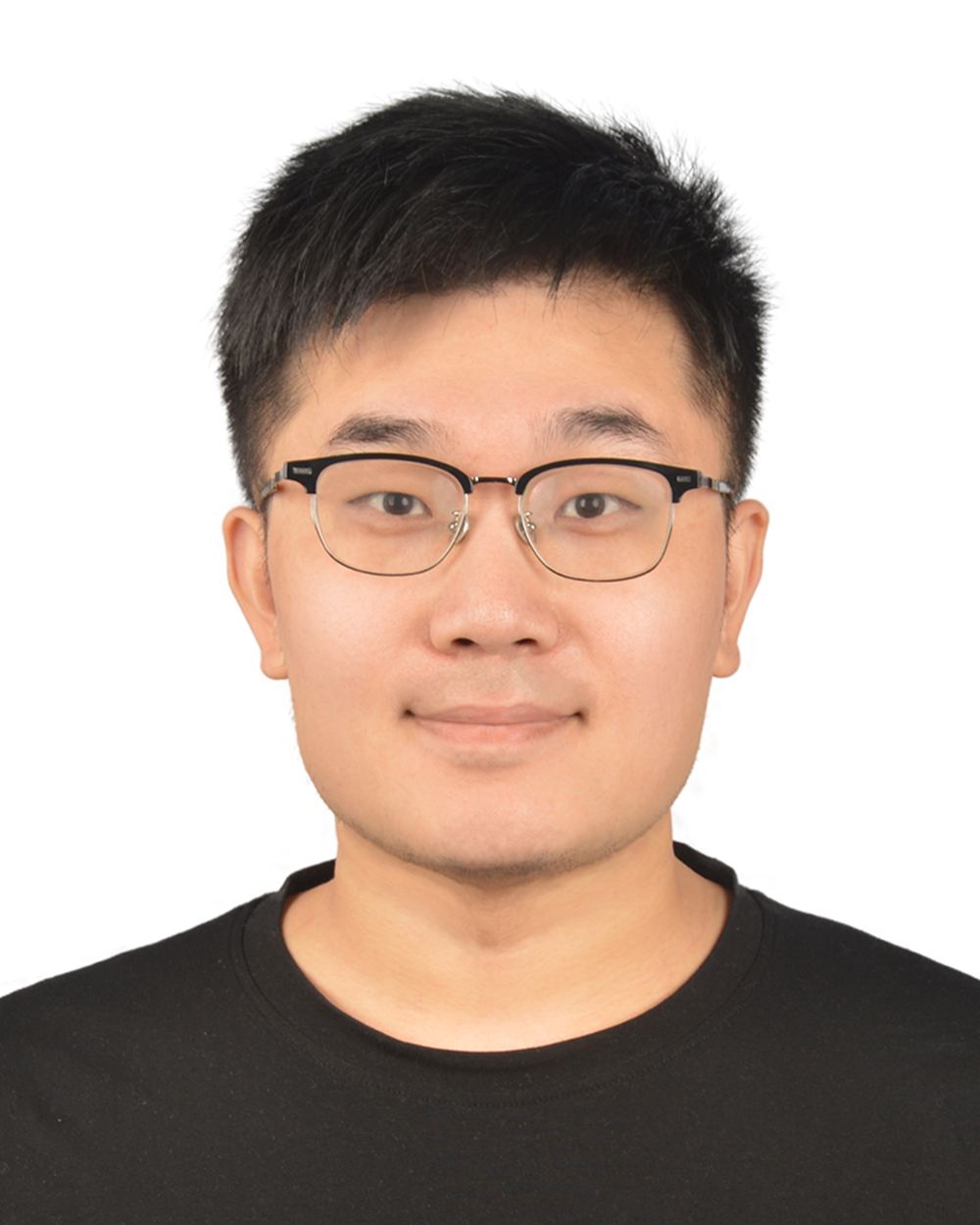}}]{Lei Luo} received the Ph.D. degree in pattern recognition and intelligence systems from the School of Computer Science and Engineering, Nanjing University of Science and Technology (NJUST), Nanjing, China. From 2017 to 2020, he was a Post-Doctoral Fellow at the University of Texas at Arlington, TX, USA, and the University of Pittsburgh, PA, USA. He is currently a Professor in the School of Computer Science and Technology of NJUST. His research interests include pattern recognition, machine learning, data mining and computer vision. Prof. Luo has served as an PC/SPC Member for IJCAI, AAAI, NeurIPS, ICML, KDD, CVPR, and ECCV, and a reviewer for over ten international journals, such as IEEE TPAMI, IEEE TIP, IEEE TSP, IEEE TCSVT, IEEE TNNLS, IEEE TKDE, and PR.	
\end{IEEEbiography}

\end{document}